\documentclass{article} 
\usepackage{iclr2025_conference,times}

\usepackage{fancyhdr}
\renewcommand{\headrulewidth}{0pt}  
\renewcommand{\footrulewidth}{0pt}  

\usepackage{amsthm}

\usepackage{amsmath,amsfonts,bm}

\def\eqref#1{equation~\ref{#1}}

\def\1{\bm{1}}

\DeclareMathAlphabet{\mathsfit}{\encodingdefault}{\sfdefault}{m}{sl}
\SetMathAlphabet{\mathsfit}{bold}{\encodingdefault}{\sfdefault}{bx}{n}

\usepackage{hyperref}
\usepackage{url}

\usepackage{multicol, multirow}
\usepackage{array}
\usepackage{tabularx}
\usepackage{booktabs}

\usepackage{caption}

\usepackage{pifont}

\usepackage{wrapfig}

\usepackage{algorithm}
\usepackage{algorithmic}

\usepackage{amsmath}
\usepackage{amssymb}
\usepackage{mathtools}
\usepackage{subfigure}

\newtheorem{theorem}{Theorem}[section]

\newtheorem{remark}[theorem]{Remark}

\usepackage{graphicx}

\title{MaD-Scientist: AI-based Scientist solving Convection-Diffusion-Reaction Equations Using Massive PINN-Based Prior Data}

\iclrfinalcopy
\author{
Mingu Kang\thanks{Equal contribution, alphabetically ordered.}
\And
Dongseok Lee\footnotemark[1]
\And
Woojin Cho
\And
Jaehyeon Park
\And
Kookjin Lee
\And
Anthony Gruber
\And
Youngjoon Hong\thanks{Co-corresponding, alphabetically ordered.}
\And
Noseong Park\footnotemark[2]
}

\begin{document}

\maketitle

\begin{abstract}
Large language models (LLMs), like ChatGPT, have shown that even trained with noisy prior data,  they can generalize effectively to new tasks through in-context learning (ICL) and pre-training techniques.
Motivated by this, we explore whether a similar approach can be applied to scientific foundation models (SFMs). Our methodology is structured as follows: (i) we collect low-cost physics-informed neural network (PINN)-based approximated prior data in the form of solutions to partial differential equations (PDEs) constructed through an arbitrary linear combination of mathematical dictionaries; (ii) we utilize Transformer architectures with self and cross-attention mechanisms to predict PDE solutions without knowledge of the governing equations in a zero-shot setting; (iii) we provide experimental evidence on the one-dimensional convection-diffusion-reaction equation, which demonstrate that pre-training remains robust even with approximated prior data, with only marginal impacts on test accuracy. Notably, this finding opens the path to pre-training SFMs with realistic, low-cost data instead of (or in conjunction with) numerical high-cost data. These results support the conjecture that SFMs can improve in a manner similar to LLMs, where fully cleaning the vast set of sentences crawled from the Internet is nearly impossible.
\end{abstract}

\section{Introduction}

In developing large-scale models, one fundamental challenge is the inherent noisiness of the data used for training. Whether dealing with natural language, scientific data, or other domains, large datasets almost inevitably contain noise. Large language models (LLMs), such as ChatGPT, present an interesting paradox: despite being trained on noisy datasets, they consistently produce remarkably clean and coherent output. This observation raises an important question for the development of scientific foundation models (SFMs): Can an SFM, like an LLM, learn from noisy data and still generate accurate, dynamic results for complex scientific tasks?

\begin{figure}[b]
  \centering
  \includegraphics[width=\textwidth]{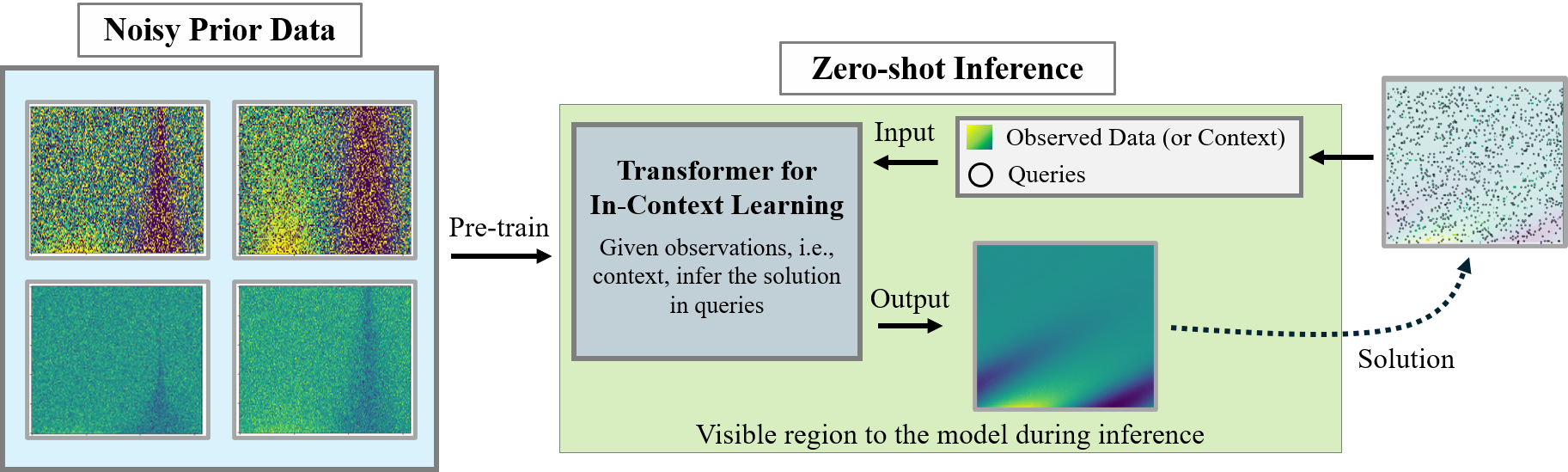}
  \caption{\textbf{An end-to-end schematic diagram of our model.} Our model performs in-context learning based on the given observations, i.e., context, to infer the solution. Even when trained with noisy PINN-prior, our model can obtain clean solutions due to its Bayesian inference capability.}
  \label{fig:end_to_end}
\end{figure}

In recent years, LLMs have revolutionized the field of natural language processing by introducing highly flexible and scalable architectures~\citep{NEURIPS2020_1457c0d6, kaplan2020scalinglawsneurallanguage, touvron2023llamaopenefficientfoundation, NEURIPS2023_58168e8a, JMLR:v24:22-1144}. Notably, the in-context learning (ICL) paradigm has demonstrated powerful generalization capabilities, enabling LLMs to adapt to new tasks without explicit fine-tuning~\citep{NEURIPS2020_1457c0d6, radford_language_2019, dai2023why, NEURIPS2023_3eb7ca52}. This success has motivated the application of such foundation models across a variety of domains~\citep{xu2024towards, xie2024llamafoundationlargelanguage, yang2024uniaudio}.
Scientific machine learning (SciML) is one such emerging domain which merges physics-based models with machine learning methodologies~\citep{RAISSI2019686,10.1145/3514228,subramanian2023towards,kim2024simple,doi:10.1021/acsmaterialslett.2c01096,CHOI2024116678}. SciML aims to leverage the power of machine learning to solve complex scientific problems, including those governed by partial differential equations (PDEs). Recent efforts in this direction have led to the development of foundation models specifically designed for scientific tasks, called SFMs~\citep{doi:10.1073/pnas.2310142120,xie2024llamafoundationlargelanguage,yang2024uniaudio,Moor2023,bodnar2024aurora,herde2024poseidon}. These models aim to generalize across a wide range of scientific problems using prior data, much like how LLMs generalize across various language tasks.
For example, the versatility of in-context operator networks (ICONs), as illustrated in studies like \citet{YANG2024113379} and \citet{doi:10.1073/pnas.2310142120}, underscores their generalization capabilities in various PDE-related tasks, particularly in the context of few-shot learning. Moreover, the integration of in-context operator learning into multi-modal frameworks, as demonstrated by ICON-LM \citep{yang2024finetunelanguagemodelsmultimodal}, has pushed the boundaries of traditional models by combining natural language with mathematical equations. Additionally, several other studies have focused on solving a family of PDEs with a single trained model~\citep{pmlr-v235-cho24b}. However, all these studies are limited in their ability to fully harness the capabilities of large foundation models.
Our methodology, called \underline{\textbf{Ma}}ssive prior \underline{\textbf{D}}ata-assisted AI-based \underline{\textbf{Scientist}} (MaD-Scientist), addresses these limitations and offers significant advantages in the following four aspects.

\paragraph{No prior knowledge of physical laws} Our goal is to predict solutions from observed quantities, such as velocity and pressure, without relying on governing equations, a common challenge in many real-world scenarios \citep{lee2024grid,nicolaou2023data,rouf2021stock, beck2021perspective,chien2012manufacturing}
. In complex systems, such as those governing semiconductor manufacturing, the exact governing equations are often unknown and may change over time \citep{chien2012manufacturing,quirk2001semiconductor}
. Therefore, excluding these equations from the model input is a strategic choice aimed at enhancing the applicability of our method across various domains.

\paragraph{Zero-shot inference} Our goal is to achieve zero-shot inference for predicting PDE solutions.
For instance, ICON-LM requires few-shot ``demos''\footnote{In ICON and ICON-LM, a demo means a set of (input, output) pairs of an operator to infer.} for an unknown target operator before making predictions. In contrast, our foundation model eliminates the need for such demos, as collecting them implies that inference cannot occur until these few-shot examples are available; see e.g., Figure~\ref{fig:end_to_end}. Our approach is designed to enable immediate inference as soon as the model is queried.

\paragraph{Bayesian inference} 
We incorporate Bayesian inference into the prediction process by leveraging prior knowledge obtained from numerical solutions in PDE dictionaries. This approach allows the model to make more accurate and well-informed predictions by defining a prior distribution over unseen PDE coefficients. During training, the model learns to capture relationships among known data points using self-attention mechanisms, while cross-attention enables it to extrapolate and infer solutions for new, unseen points. When tested, the model utilizes this prior data to generalize effectively to novel data points, achieving zero-shot predictions without the need for additional fine-tuning.

\paragraph{Approximated prior data} For LLMs, one of the most challenging steps is collecting prior data, which typically involves crawling and cleaning sentences from the Internet. However, this process is far from perfect due to two key issues: (i) the Internet, as a data source, is inherently unreliable;  (ii) cleaning such vast amounts of data requires significant manual effort. Consequently, LLMs are often trained on incomplete or imperfect prior data. Remarkably, this realistic yet critical issue has been largely overlooked in the literature on SFMs, despite their similarities to LLMs. For example, when generating data using numerical solvers for PDEs without known analytical solutions, numerical errors inevitably arise, manifesting as a form of measurement noise 
--- for this, we conducted preliminary experiments for training SFMs with noisy data in Appendix~\ref{a:noisy}, which shows the possibility of training SFMs with data inexact to some degree. 

Moreover, numerical solvers running on large-scale servers are frequently expensive and time-consuming and they are typically optimized towards certain types of PDEs, e.g., the finite-difference time-domain (FDTD) method for Maxwell's equations. In this work, we are the first to explore the potential of pre-training SFMs with PINN-based low-cost/noisy/approximated data.

\paragraph{CDR Study} For our empirical studies, we use a family of the convection-diffusion-reaction (CDR) equation with various types of reaction, which serves as a paradigm problem representing generic elliptic equations. By solving the CDR equation, our approach can be extended to a wide range of other problems. We compare our method with two state-of-the-art machine learning techniques for solving parameterized PDEs. Additionally, we introduce three different types of noise into the numerical solutions of the CDR equation. Our approach not only outperforms the two baseline methods but also demonstrates stable performance, even when noise is added to the prior data during pre-training.

\section{Background}

Consider a sequence of pairs $(X_{1},Y_{1}), (X_{2},Y_{2}), \ldots$, each within the measurable space $(\mathcal{X} \times \mathcal{Y}, \mathcal{A})$, where \(X_i\) represents the spatiotemporal coordinate, \(Y_i\) denotes the corresponding solution in this paper's context and $\mathcal{A}$ denotes the Borel $\sigma$-algebra on the measurable space $\mathcal{X} \times \mathcal{Y}$. For simplicity, we adopt this notation in this section. These pairs are drawn from a family of probability density distributions \(\{p_q : q \in \mathcal{Q}\}\), commonly referred to as the \textit{statistical model}, where \(\mathcal{Q}\) represents the \textit{parameter space} equipped with a \(\sigma\)-algebra \(\mathcal{B}\) ensuring that the mappings \(q \mapsto p_q(x, y)\) are measurable. The true underlying density function \(\pi\) is a member of $\mathcal{Q}$, and the pairs \((X_i, Y_i)\) are sampled according to \(p_\pi\).
Lacking information about $\pi$, we adopt a Bayesian framework to establish a prior distribution $\Pi$ which is defined as probability measure on \( (\mathcal{Q}, \mathcal{B}) \). Then we have, for any measurable set $A \in \mathcal{B}$,
\begin{equation}
\Pi(A \mid X,Y) = \frac{\int_A p_q(X,Y) \mathrm{d}\Pi(q)}{\int_\mathcal{Q} p_q(X,Y)\mathrm{d}\Pi(q)}.
\end{equation}
Let us adopt the notation $p_q=q$. This prior is updated with the observed data to form the posterior distribution, which is defined as
\begin{equation}
\Pi(A \mid D_n ) = \frac{\int_A L_n(q) \mathrm{d}\Pi(q)}{\int_\mathcal{Q} L_n(q) \mathrm{d}\Pi(q)},
\end{equation}
where $L_n(q) = \prod_{i=1}^n \frac{q(X_{i},Y_{i})}{\pi(X_{i},Y_{i})}$ for $A \subset \mathcal{Q}$ and \( D_n = \{(X_i, Y_i)\}_{i=1}^{n} \). 
The resulting posterior density is
\begin{equation}
q_n(X,Y \mid D_n) = \int_\mathcal{Q} q(X,Y) \mathrm{d}\Pi(q \mid D_n),
\end{equation}
 and the posterior predictive distribution (PPD) is formulated as
\begin{equation}
\pi(y \mid x, D_n) = \int_\mathcal{Q} q(y \mid x) \, \mathrm{d}\Pi(q \mid D_n).
\end{equation}
The behavior of \(D_n\) plays a crucial role in this formulation. 
As noted by \citet{walker2004modern,walker2004new,6e9902fe-c806-359d-8aa7-a66426017a1c,walker2003sufficient,nagler2023statisticalfoundationspriordatafitted}, for a well-behaved prior, the PPD converges toward \(\pi\) as \(n\) increases. This aligns with findings in \citet{6e9902fe-c806-359d-8aa7-a66426017a1c}, demonstrating that in well-specified scenarios, strong consistency is achieved as
\begin{equation}\label{eq:consistency}
\Pi^n \left(\{ q : H(\pi, q) > \epsilon \}\right) \to 0 \quad \text{almost surely}, 
\end{equation}
for any \(\epsilon > 0\),
where \(\Pi^n(A) = \int_A d\Pi(q \mid D_n)\) is the posterior measure and \(H\) is the Hellinger distance defined by
\[
H(p, q) = \left(\int_{\mathcal{X}\times\mathcal{Y}} (\sqrt{p} - \sqrt{q})^2 \,\right)^{1/2}.
\] 

\begin{theorem}\label{neural_sensitivity}
Suppose that for any \( \epsilon > 0 \), there exists a Transformer parameterized by \(\hat{\theta}\) such that 
\[
\hat{\theta} = \underset{\theta}{\arg\min} \; \mathbb{E}_{x} \left[ KL \left( p_{\theta}(\cdot \mid x, D_n), \pi(\cdot \mid x, D_n) \right) \right] < \epsilon,
\] for any realization of  \(D_n\).
If the posterior consistency condition \eqref{eq:consistency} holds, and for any \( q \in \mathcal{Q} \), \( q(x) = \pi(x) \) almost everywhere on \( \mathcal{X} \), then the following holds almost surely (see Appendix~\ref{a:proof} for proof):
\[
\mathbb{E}_{x} \left[ H \left( p_{\hat{\theta}}(\cdot \mid x, D_n), \pi(\cdot \mid x) \right) \right] \xrightarrow{n \to \infty} 0.
\]
\end{theorem}

This result demonstrates that as the amount of data increases, the neural network close to  the posterior distribution converges to the expected value under the prior distribution, highlighting the consistency and robustness of the Bayesian inference process. Based on this observation, our model performs the Bayesian inference with prior data. Ultimately, our model's goal is to read some ground-truth spatiotemporal points and infer an appropriate PDE solution that accurately describes the dynamics under the given spatiotemporal conditions.

\section{Methods}

Suppose the dataset \( D_n = \{(X_i,T_i, Y_i)\}_{i=1}^{n} \) is independently and identically distributed (i.i.d.) and sampled from a distribution \( q_{\alpha} \), where \( \alpha \) is the parameter vector representing the coefficients governing the PDE dynamics, including convection, diffusion, and reaction terms. Specifically, \( Y_i \sim u(X_i, T_i \mid \alpha) + \text{noise} \), where the noise represents the difference between the PINN-predicted solution \(\Tilde{u}(\alpha)\) and the true solution \(u(\alpha)\). The PPD of the solutions given the dataset can be expressed as 
\begin{equation} q(y \mid x,t, D_n) = \int_H q_{\alpha}(y \mid x,t) \, \mathrm d\Pi(q_{\alpha} \mid D_n), \end{equation} 
which represents the likelihood distribution of \(y\) given \(D_n\), capturing the most probable solution distribution for the given parameter $\alpha$. In this work, we aim to predict the solution from \( D_n \) by minimizing the mean squared error (MSE) between the PPD-derived solution and the true solution, even in the presence of noise. This requires constructing a prior over the PDE solution space, which is detailed next.

\paragraph{Benchmark PDE} The following one-dimensional convection-diffusion-reaction (CDR) equation is used for the benchmark PDE, 
\begin{equation}
\begin{aligned}
\label{eqs:1D_CDR}
\textnormal{1D CDR: } u_t + \beta u_x - \nu u_{xx} - \rho f(u) = 0, \;\;\;\; x \in [0, 2\pi], t \in [0, 1],
\end{aligned}
\end{equation}
where $f:\mathbb{R} \rightarrow \mathbb{R}$ is a reaction term such as Fisher, Allen-Cahn and Zeldovich. This equation consists of three key terms with distinct properties: convective, diffusive, and reactive, making it an ideal benchmark problem. It is commonly used in the PINN literature due to the diverse dynamics introduced by its three parameters: $\beta, \nu,$ and $\rho$, which include various failure modes~\citep{krishnapriyan2021characterizing}. 
To our knowledge, however, our work is the first predicting all those different reaction terms with a single model.

In this paper, we use the following dictionary of CDR-related terms, incorporating a linear combination of $J$ nonlinear reaction terms, for generating prior data.
\begin{equation}
\label{eqs:governing equation}
\begin{aligned}
    u_t = \mathcal{N}(\cdot), \;\;\;\; & \mathcal{N}(t, x, u, \beta, \nu, \rho_1, \cdots, \rho_J) = -\beta u_x + \nu u_{xx} + \sum_{j=1}^{J} \rho_{j} f_j(u),
\end{aligned}
\end{equation}
where each $f_j$ represents specific reaction term. This expansion allows for the introduction of diverse reaction dynamics. 
One can solve CDR equations with numerical solvers. In this work, however, we are interested in building low-cost PINN-based prior data. In the future, one may need to build prior data for not only CDR but also many other equations for which none of analytical/numerical solutions are obtainable in a low-cost manner, e.g., Naiver-Stokes equations. We think our PINN-based prior data will play a crucial role in such a case.

\paragraph{PINN-Prior of PDE Solution Space}
To approximate the solution space for PDEs,
we construct a parameter space, $\Omega$, which is the collection of coefficients in~\eqref{eqs:governing equation}:
\begin{equation}
\label{eqs:parameter space}
\begin{aligned}
    \Omega = \{\alpha &:= (\beta, \nu, \rho_1, \cdots, \rho_{J}) \},
\end{aligned}
\end{equation}
which has a dictionary form. Consequently, the target exact prior $\mathcal{U}$ represents the collection of solutions $u(\alpha)$ at \eqref{eqs:governing equation} for each parameter $\alpha \in \Omega$, where $\mathcal{X}$ and $\mathcal{T}$ correspond to the spatial and temporal domains of interest, respectively
\begin{equation}
\label{eqs: target prior}
\begin{aligned}
\mathcal{U} = \bigcup_{\alpha \in \Omega} \{ u(\alpha) \,|\, u_t = \mathcal{N}(t, x, u, \alpha) \}, \;\;\;\; \mathcal{U}: \mathcal{X} \times \mathcal{T} \rightarrow \mathbb{R}.
\end{aligned}
\end{equation}

 Since the target exact prior data $\mathcal{U}$ is hard to obtain, we instead use a PINN-prior $\mathcal{D}$ that closely approximates $\mathcal{U}$ as follows. Suppose $\Tilde{u}(\alpha)$ is the prediction by PINN (Appendix \ref{a:PINN}) which is trained to predict the PDE $u_t = \mathcal{N}(\cdot)$. The PINN-prior $\mathcal{D}$ is a collection of approximated solutions $\Tilde{u}(\alpha)$ for each $\alpha \in \Omega$,
\begin{equation}
\label{eqs: approximating prior}
\begin{aligned}
\mathcal{D} = \bigcup_{\alpha \in \Omega} \{ \Tilde{u}(\alpha) \}, \;\;\;\; p(\mathcal{D}) \sim p(\mathcal{U}).
\end{aligned}
\end{equation}
Subsequently, the model learns the PPD of the generated prior $p(\mathcal{D})$ through ICL.

\begin{figure}[t]
  \centering
  \includegraphics[width=0.9\textwidth]{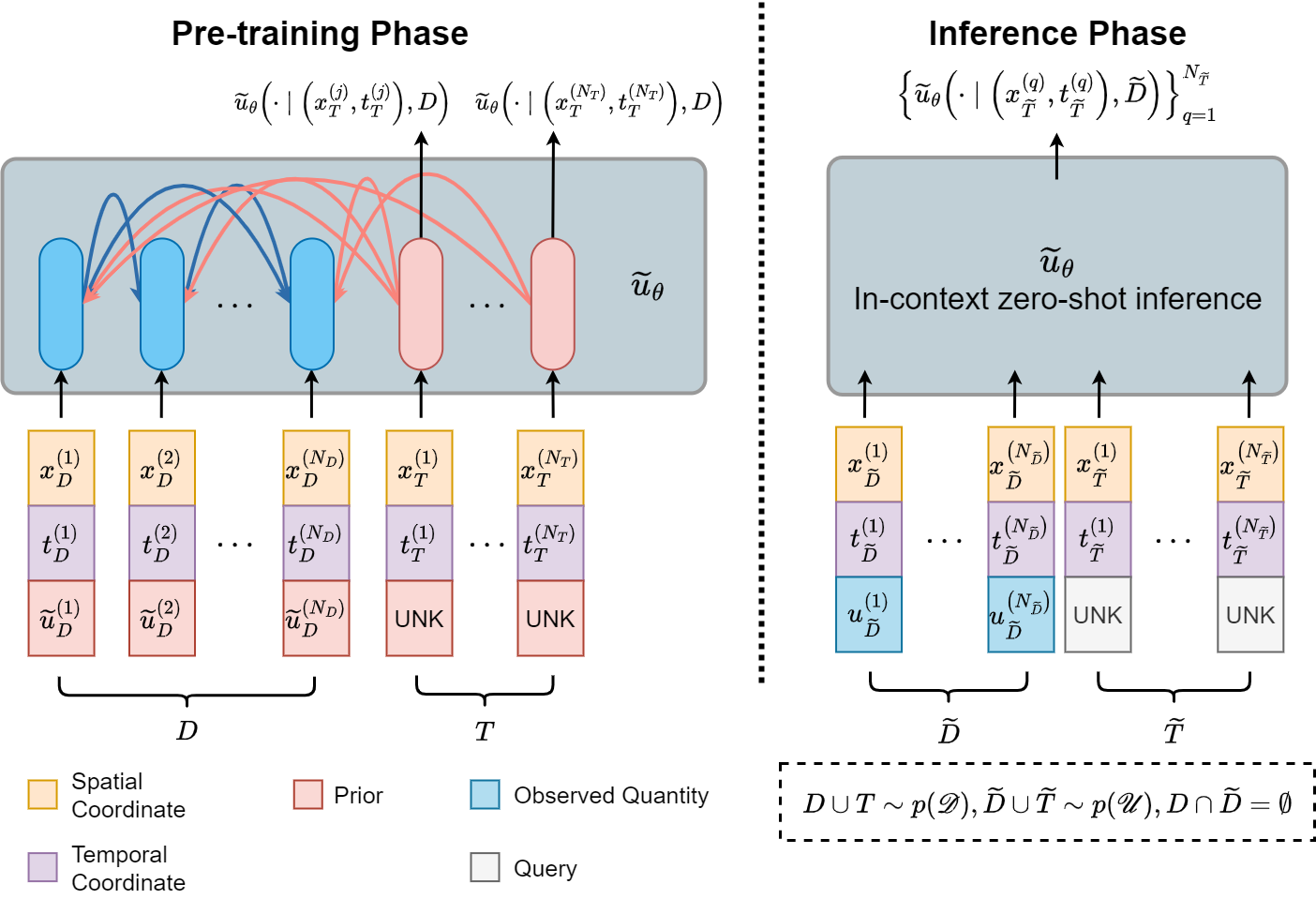}
  \caption{\textbf{A schematic diagram of Transformer.} \textit{(Left)} The Transformer $\Tilde{u}_{\theta}$ takes prior of solution-known $D$ and querying task $T$ drawn from the prior distribution $\mathcal{D}$ and infers solutions of the queried points in the training phase. ICL is leveraged with a self-attention among $D$ (blue rods) and a cross-attention from $T$ to $D$ (red rods). \textit{(Right)} In the testing phase, $\Tilde{u}_{\theta}$ takes an input of unseen data $\widetilde{D}$ and $\widetilde{T}$ drawn from the ground truth distribution $\mathcal{U}$, and the model predicts the queried points $\widetilde{T}$.}
  \label{fig:Train_Transformer}
\end{figure}

\paragraph{Training}\label{training} From a given parameter space $\Omega$, the parameter $\alpha$ is randomly drawn i.i.d. from $\Omega$. This method is adopted from meta learning \citep{DBLP:conf/icml/FinnAL17} which optimizes the model parameter to adapt to various tasks, in our case the prediction over wide prior space $\mathcal{D}$ expressed as a dictionary over $\alpha$. After that, the previous $\Tilde{u}(\alpha)$ is then given as an input to Transformer $\Tilde{u}_{\theta}$ to minimize the mean square error (MSE) at the predicted points, see  \eqref{eqs:train_loss}. The MSE loss criterion is proposed as the Transformer's task is to perform regression of the solution over the spatial and temporal domain for given $\Tilde{u}(\alpha)$,
\begin{equation}
L_{\alpha} = \frac{1}{N_T} \sum_{j=1}^{N_T} \left[ \Tilde{u}_{\theta}(x_T^{(j)}, t_T^{(j)} \mid D_n) - \Tilde{u}(x_T^{(j)}, t_T^{(j)})\right]^2.
\label{eqs:train_loss}
\end{equation}

\paragraph{Evaluation}
To illustrate the model's zero-shot learning capability in scenarios commonly encountered in practical applications, we assess the model's performance using data $\Tilde{D}$ sampled i.i.d. from $\mathcal{U}$, not overlapping with the training set $D \cup T$. For evaluation, we employ both absolute and relative $L_2$ errors between the model's predicted solutions for test queries and the numerically computed ground truth. These errors are then averaged over the target parameter space $\Omega$ used during training.


\section{Experiments}
\label{S:Experiments}


Our experiment section is divided into two phases: in the first phase, we conduct a focused study with the basic reaction term, Fisher, to understand the base characteristics of SFMs, and in the second phase we conduct comprehensive studies with various reaction terms.

\subsection{Experimental Setup}
\label{ss:Experimental_setup}

\paragraph{Baseline methods} We compare our model with 2 baselines: Hyper-LR-PINN \citep{NEURIPS2023_24f8dd1b} 
and P$^2$INN without fine tuning~\citep{pmlr-v235-cho24b}. Both models are parametrized PINNs designed to learn parameterized PDEs. Hyper-LR-PINN emphasizes a low-rank architecture with a parameter hypernetwork, while P$^2$INN focuses on a parameter-encoding scheme based on the latent space of the parameterized PDEs.

Following this, as shown in Figure \ref{fig:Train_Transformer}, the model takes $D \cup T \sim p(\mathcal{D})$ in training phase and $\widetilde{D} \cup \widetilde{T} \sim p(\mathcal{U})$ in testing phase. In addition, the dataset $D \cup T$ requires the prior $\widetilde{u}$, and $\widetilde{D}$ requires the solution $u$. For a fair comparison, we use $D$, $T$, and $\widetilde{D}$ as the training dataset for both Hyper-LR-PINN and P$^2$INN. Notably, while Hyper-LR-PINN and P$^2$INN do not rely on solution points during training and testing, our model operates without any knowledge of the governing equation $\mathcal{N}(\cdot)$. This setup ensures a valid and balanced comparison (Table \ref{table:baseline_comparison}). The additional comparison details are elaborated in Appendix \ref{a:Comparison}, providing further insights into the distinctions between these models.


\begin{table}
\caption{Major comparisons between Hyper-LR-PINN, P$^2$INN, and our model. While both Hyper-LR-PINN and P$^2$INN require the knowledge of governing equation, our model only needs observed quantities. The notations used in the table are fully aligned with those in Figure \ref{fig:Train_Transformer}.}
\label{table:baseline_comparison}
\small
\centering
\begin{tabularx}{\textwidth}{lXXXr}
\toprule
\textbf{Properties} & \textbf{Hyper-LR-PINN} & \textbf{P$^2$INN} & \textbf{Ours} \\
\midrule
Target function & $u(x, t; \alpha \in \Omega)$ & $u(x, t; \alpha \in \Omega)$ & $u(x, t)|_{\mathcal{D}}$\\
Governing equation $\mathcal{N}(\cdot)$ given & $\checkmark$ & $\checkmark$ & \ding{55}\\
Train dataset & $D \cup T \cup \widetilde{D}$ & $D \cup T \cup \widetilde{D}$ & $D \cup T$ \\
Test dataset & $\widetilde{T}$ & $\widetilde{T}$ & $\widetilde{D} \cup \widetilde{T}$ \\
Dataset with a solution & None & None & $D, T, \widetilde{D}$\\
\bottomrule
\end{tabularx}
\end{table}

\paragraph{Training algorithm} The concrete flow of training phase is described in Algorithm at Appendix \ref{a: Training Algorithm}.

\subsection{Focused Study to Better Understand SFMs' base characteristics}
\label{SS:focused_study}

In this section, we employ six different dynamics derived from the 1D CDR equation with a Fisher reaction term, \( u_t + \beta u_x - \nu u_{xx} - \rho u(1-u) = 0 \) (Appendix \ref{a:datasets}). We begin with an in-depth study using the Fisher reaction term, chosen for its simplicity among the reaction terms, which has been extensively studied in population dynamics \citep{ALKHALED2001245}. This allows us to better understand the core characteristics of the SFM, facilitating a more effective analysis of the model’s behavior.
\subsubsection{Time Domain Interpolation for Seen PDE Parameters with a Numerical Prior}
\label{SSS:time_inter_numerical}

We first verify the ICL capability of Transformer with a numerical prior, i.e., $\Tilde{u}(\alpha)$ equals to the solution $u$ of the PDE $u_t = \mathcal{N}(t, u, x, \alpha)$, before we dive into using a PINN prior. For each equation, we set the parameter space $\Omega$ with three different coefficient ($\beta, \nu, \rho$) range: $\left([1, 5] \cap \mathbb{Z} \right)^{m}, \left([1, 10] \cap \mathbb{Z} \right)^{m}$, and $\left([1, 20] \cap \mathbb{Z} \right)^{m}$, where $m$ is the number of nonzero coefficients. The Transformer $\widetilde{u}_{\theta}$ is trained with $D \cup T \subseteq u(\alpha)$ where $\alpha \in \Omega$ is selected uniformly at random for each epoch. After that, we test $\widetilde{u}_{\theta}$ with $\widetilde{D} \cup \widetilde{T} \subseteq u(\alpha)$ for all $\alpha \in \Omega$ and evaluate average $L_2$ absolute and relative error (Table \ref{table:5.1.1}).

We highlight two key observations: First, our model outperforms baseline models applied to diffusion, reaction, reaction-diffusion, and convection-diffusion-reaction systems. Second, it demonstrates stable performance across a wide range of coefficient values.
For instance, all baselines show difficulties in predicting accurate solutions for high coefficients, especially in diffusion and reaction systems, while ours do not. When we measure the standard deviation of $L_2$ relative error over three coefficient range for diffusion system, ours have $9.1 \times 10^{-4}$ while others show $10^{-2}$ scale value. These observations not only verify the effectiveness of the Transformer's ICL capability, but also suggest its potential to handle larger parameter space $\Omega$.

\begin{table*}[t]
\caption{The relative and absolute $L_2$ errors over the 1D-CDR equation using a numerical prior. P$^2$INN is tested without fine-tuning, and *-marked cases are evaluated with a reduced number of parameters due to the extensive computational requirements.}
\begin{center}
\label{table:5.1.1}
\makebox[\textwidth][c]{
\resizebox{0.65\paperwidth}{!}{%
\begin{tabular}{lccccccccr}
\toprule
\multirow{2}{*}{\textbf{System}} & \multirow{2}{*}{\textbf{Coefficient range}} & \multicolumn{2}{c}{\textbf{Hyper-LR-PINN}} & \multicolumn{2}{c}{\textbf{P$^2$INN} } & \multicolumn{2}{c}{\textbf{Ours}} \\
&& Abs.err & Rel.err & Abs.err & Rel.err & Abs.err & Rel.err \\
\midrule

\multirow{3}{*}{\textbf{Convection}} 
& $\beta \in [1, 5] \cap \mathbb{Z}$ & \textbf{0.0104} & \textbf{0.0119} & 0.0741 & 0.1020 & 0.0192 & 0.0184\\ 
& $\beta \in [1, 10] \cap \mathbb{Z}$ & \textbf{0.0172} & \textbf{0.0189} & 0.1636 & 0.1801 & 0.0250 & 0.0251\\ 
& $\beta \in [1, 20] \cap \mathbb{Z}$ & \textbf{0.0340} & \textbf{0.0368} & 0.2742 & 0.2743 & 0.0764 & 0.0864\\
\cmidrule(lr){1-8}

\multirow{3}{*}{\textbf{Diffusion}} 
& $\nu \in [1, 5] \cap \mathbb{Z}$ & 0.0429 & 0.0570 & 0.3201 & 0.3652 & \textbf{0.0096} & \textbf{0.0120}\\ 
& $\nu \in [1, 10] \cap \mathbb{Z}$ & 0.0220 & 0.0282 & 0.3550 & 0.4029 & \textbf{0.0108} & \textbf{0.0137}\\ 
& $\nu \in [1, 20] \cap \mathbb{Z}$ & 0.1722 & 0.1991 & 0.4553 & 0.5166 & \textbf{0.0095} & \textbf{0.0134}\\ 
\cmidrule(lr){1-8}

\multirow{3}{*}{\textbf{Reaction}} 
& $\rho \in [1, 5] \cap \mathbb{Z}$ & 0.0124 & 0.0428 & 0.0109 & 0.0354 & \textbf{0.0102} & \textbf{0.0154}\\ 
& $\rho \in [1, 10] \cap \mathbb{Z}$ & 0.2955 & 0.3562 & 0.0192 & 0.0708 & \textbf{0.0129} & \textbf{0.0202}\\ 
& $\rho \in [1, 20] \cap \mathbb{Z}$ & 0.7111 & 0.7650 & 0.1490 & 0.2915 & \textbf{0.0160} & \textbf{0.0322}\\ 

\cmidrule(lr){1-8}
\multirow{3}{*}{\textbf{Convection-Diffusion}} 
& $\beta, \nu \in [1, 5] \cap \mathbb{Z}$ & \textbf{0.0046} & \textbf{0.0055} & 0.1329 & 0.1554 & 0.0195 & 0.0231\\ 
& $\beta, \nu \in [1, 10] \cap \mathbb{Z}$ & 0.0268 & 0.0295 & 0.1609 & 0.1815 & \textbf{0.0211} & \textbf{0.0274}\\ 
& $\beta, \nu \in [1, 20] \cap \mathbb{Z}$ & *0.1487 & *0.1629 & 0.1892 & 0.2044 & \textbf{0.0226} & \textbf{0.0305}\\ 

\cmidrule(lr){1-8}
\multirow{3}{*}{\textbf{Reaction-Diffusion}} 
& $\nu, \rho \in [1, 5] \cap \mathbb{Z}$ & 0.0817 & 0.1160 & 0.0579 & 0.1346 & \textbf{0.0139} & \textbf{0.0189}\\ 
& $\nu, \rho \in [1, 10] \cap \mathbb{Z}$ & 0.0317 & 0.0446 & 0.4398 & 0.5457 & \textbf{0.0122} & \textbf{0.0189}\\ 
& $\nu, \rho \in [1, 20] \cap \mathbb{Z}$ & *0.3228 & *0.3844 & 0.1513 & 0.2955 & \textbf{0.0165} & \textbf{0.0331}\\ 

\cmidrule(lr){1-8}
\multirow{3}{*}{\textbf{Convection-Diffusion-Reaction}} 
& $\beta, \nu, \rho \in [1, 5] \cap \mathbb{Z}$ & 0.0231 & 0.0307 & 0.0418 & 0.0595 & \textbf{0.0143} & \textbf{0.0209}\\ 
& $\beta, \nu, \rho \in [1, 10] \cap \mathbb{Z}$ & *0.3135 & *0.3732 & 0.0367 & 0.0624 & \textbf{0.0276} & \textbf{0.0411}\\ 
& $\beta, \nu, \rho \in [1, 20] \cap \mathbb{Z}$ & *0.9775 & *0.9958 & 0.0446 & 0.1211 & \textbf{0.0159} & \textbf{0.0310}\\ 

\cmidrule(lr){1-8}
\multirow{2}{*}{\textbf{Statistics}} & Average & 0.1805 & 0.2033 & 0.1709 & 0.2222 & \textbf{0.0196} & \textbf{0.0267}\\

& Standard Deviation & 0.2581 & 0.2727 & 0.1423 & 0.1549 & \textbf{0.0147} & \textbf{0.0164}\\

\bottomrule
\end{tabular}
}}
\end{center}
\vskip -0.2in
\end{table*}

\subsubsection{Time Domain Interpolation for Seen PDE Parameters with a PINN-Priors}
\label{SSS:time_inter_pinn}
The Transformer has demonstrated strong ICL capabilities when trained with numerical priors. Our main focus now is to determine if this same success can be achieved using a PINN-prior. As outlined in Appendix~\ref{a:Comparison}, our preliminary results show that the Transformer remains robust even when numerical priors are subject to  various types of noise. 
Building on this, we examine how the model performs when mixing low-cost PINN-priors with numerical priors in different proportions, assessing its stability and robustness when incorporating PINN-priors.

Specifically, we train the model using the convection, diffusion, and Fisher reaction equations with integer coefficients ranging from 1 to 20. For each equation, we evaluate the model with a prior that is a mixture of PINN-prior and numerical prior in varying ratios: 0\%, 20\%, 40\%, 60\%, 80\%, and 100\% PINN-priors. Table \ref{table:PINN_Prior} indicates the absolute and relative errors $L_2$ for each setup compared to the baseline results in Section~\ref{SS:focused_study}. Furthermore, the average $L_2$ error of the PINN-prior, compared to the numerical solution, is presented to demonstrate the quality of the PINN-prior.

As a result, mixing PINN-priors with numerical priors does not significantly impact performance, as the $L_2$ absolute and relative errors remain consistent with other baselines. This indicates that a Transformer can maintain ICL capability even when trained with PINN-prior data. 
{
Also, this finding confirms that the model can effectively infer solutions from limited observed data \(\Tilde{D}\), even in the presence of inaccurate PINN-priors. 

\begin{remark}
Notably, in diffusion case, the prior exhibits an error of \(14\%\), yet our prediction error stands at just \(2.6\%\). This discrepancy not only highlights the Transformer model's strong ICL capability but also demonstrates a form of superconvergence, where the model significantly outperforms expectations given the inaccurate prior. Such a result underscores the robustness and adaptability of our approach, reinforcing the idea that even with flawed prior information, the Transformer can extract meaningful insights and achieve high accuracy in predictions.

\end{remark}
}

\begin{table*}[t]
\captionsetup{skip=1.5pt}
\caption{$L_2$ error for convection, diffusion, and reaction equations measured at seen parameters, where the parameter values range from 1 to 20. Results are provided for each PINN-Prior Ratio, and for comparison, the results of baseline models are also included.}
\vskip 0.15in
\begin{center}
\label{table:PINN_Prior}
\makebox[\textwidth][c]{
\resizebox{0.6\paperwidth}{!}{%
\begin{tabular}{lcccccc}
\toprule

\multicolumn{2}{c}{\textbf{Model}} & \parbox[c]{2.3cm}{\centering $\mathbf{L_2}$ \textbf{Error} \\ \textbf{Metric}} & 
\parbox[c]{2.3cm}{\centering \textbf{convection} \\ $\beta \in [1, 20] \cap \mathbb{Z}$}  & \parbox[c]{2.3cm}{\centering \textbf{diffusion} \\ $\nu \in [1, 20] \cap \mathbb{Z}$} & \parbox[c]{2.3cm}{\centering \textbf{reaction} \\ $\rho \in [1, 20] \cap \mathbb{Z}$} \\

\midrule
\multirow{14}{*}{\textbf{Ours}} &
\multirow{2}{*}{\textbf{0\%}} & Abs. err & 0.0764 & 0.0095 & 0.0160\\ 
&& Rel. err & 0.0864 & 0.0134 & 0.0322\\ 

\cmidrule(lr){2-6}

&\multirow{2}{*}{\textbf{20\%}} & Abs. err & 0.1197 & 0.0088 & 0.0169\\ 
&& Rel. err & 0.1276 & 0.0148 & 0.0390\\ 

\cmidrule(lr){2-6}

&\multirow{2}{*}{\textbf{40\%}} & Abs. err & 0.1543 & 0.0103 & 0.0286\\ 
&& Rel. err & 0.1582 & 0.0149 & 0.0677 \\ 

\cmidrule(lr){2-6}

&\multirow{2}{*}{\textbf{60\%}} & Abs. err & 0.1743 & 0.0172 & 0.0267\\ 
&& Rel. err & 0.1746 & 0.0208 & 0.0679\\ 

\cmidrule(lr){2-6}

&\multirow{2}{*}{\textbf{80\%}} & Abs. err & 0.1677 & 0.0217 & 0.0327\\  
&& Rel. err & 0.1713 & 0.0300 & 0.0970\\ 

\cmidrule(lr){2-6}

&\multirow{2}{*}{\textbf{100\%}} & Abs. err & 0.1563 & 0.0200 & 0.0362\\ 
&& Rel. err & 0.1654 & 0.0265 & 0.1136\\ 

\cmidrule(lr){2-6}

&\multirow{2}{*}{\textbf{Prior Loss}} & Abs. err & 0.0439 & 0.1262 & 0.0215\\ 
&& Rel. err & 0.0441 & 0.1444 & 0.0845\\ 

\cmidrule(lr){1-6}

\multicolumn{2}{c}{\multirow{2}{*}{\textbf{Hyper-LR-PINN}}} & Abs. err & 0.0340 & 0.1722 & 0.7111\\ 
&& Rel. err & 0.0368 & 0.1991 & 0.7650\\ 

\cmidrule(lr){1-6}

\multicolumn{2}{c}{\multirow{2}{*}{\textbf{P$^2$INN}}} &  Abs. err & 0.2742 & 0.4553 & 0.1490\\ 
&& Rel. err & 0.2743 & 0.5166 & 0.2915\\ 



\bottomrule
\end{tabular}
}}
\end{center}
\vskip -0.1in
\end{table*}

\subsubsection{Time Domain Interpolation for Unseen PDE Parameters}
\label{SSS:time_inter_parameter}

\begin{figure}[t]
\centering
\includegraphics[width=0.8\columnwidth]{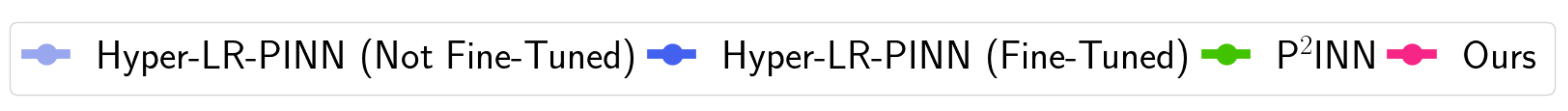}\\
\subfigure[convection]{\includegraphics[width=0.31\columnwidth]{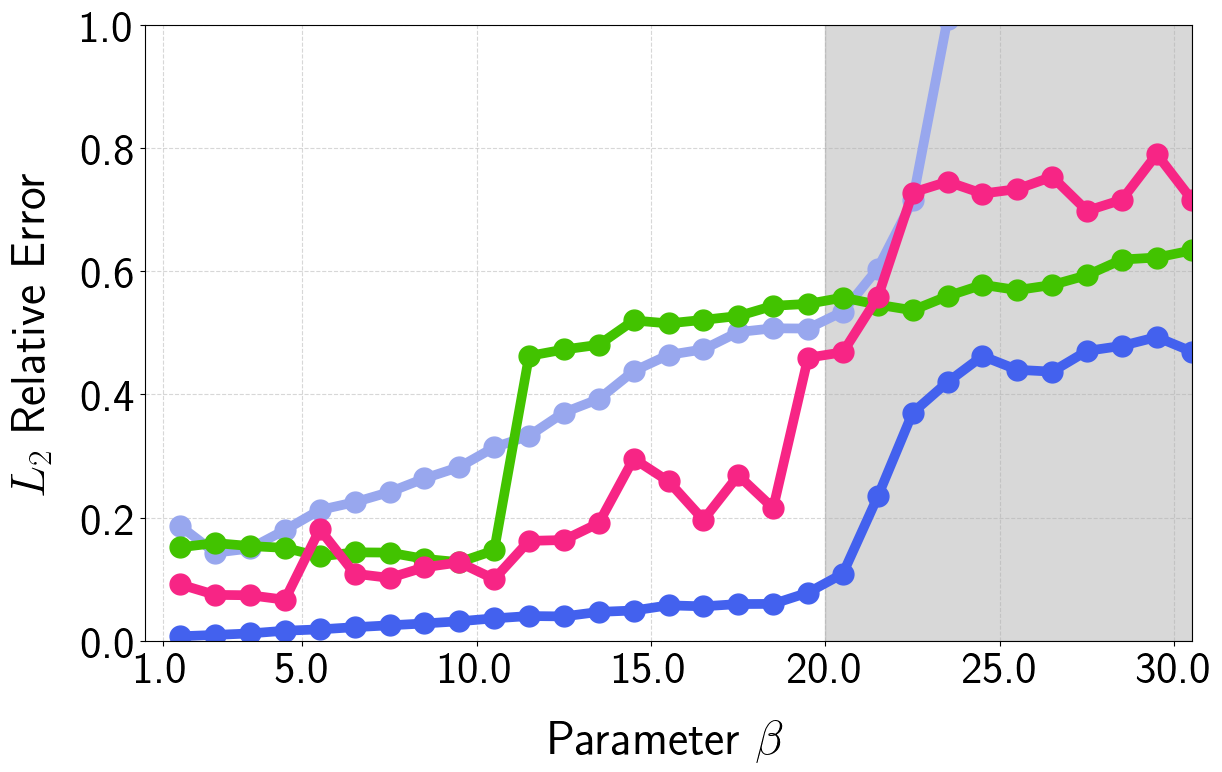}} \hfill
\subfigure[diffusion]{\includegraphics[width=0.31\columnwidth]{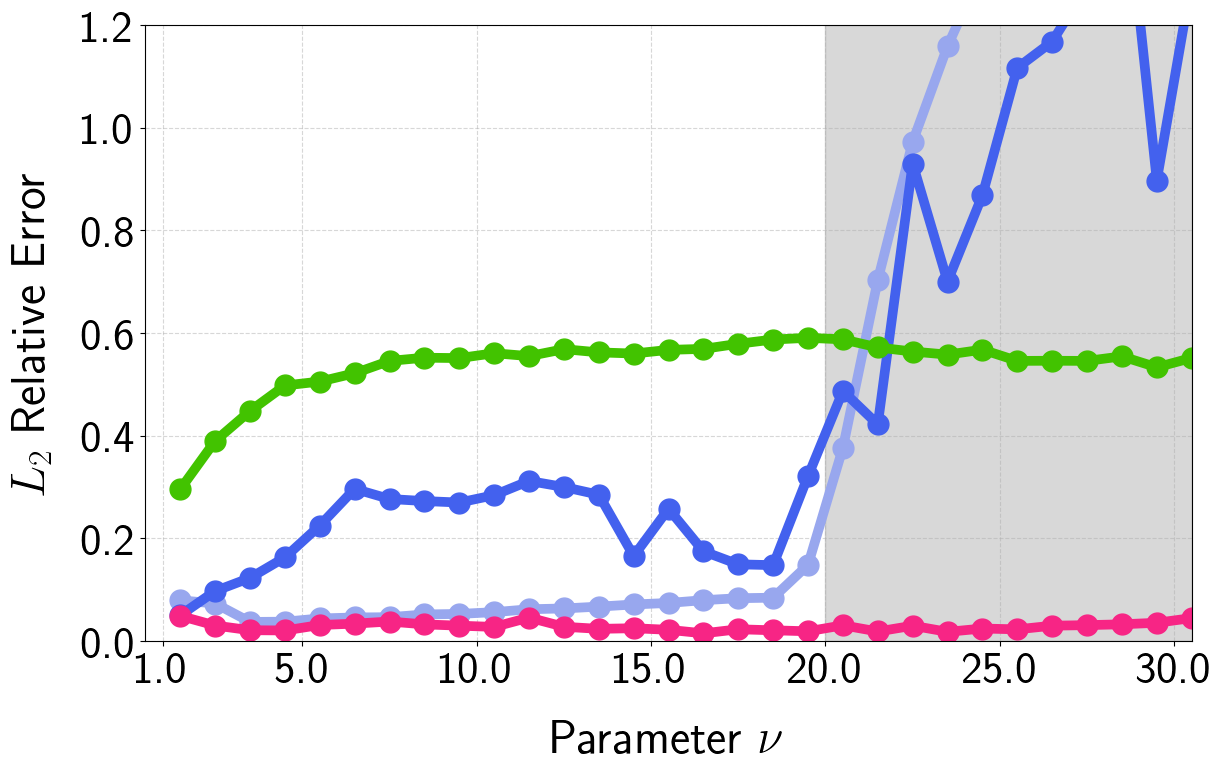}} \hfill
\subfigure[reaction]{\includegraphics[width=0.31\columnwidth]{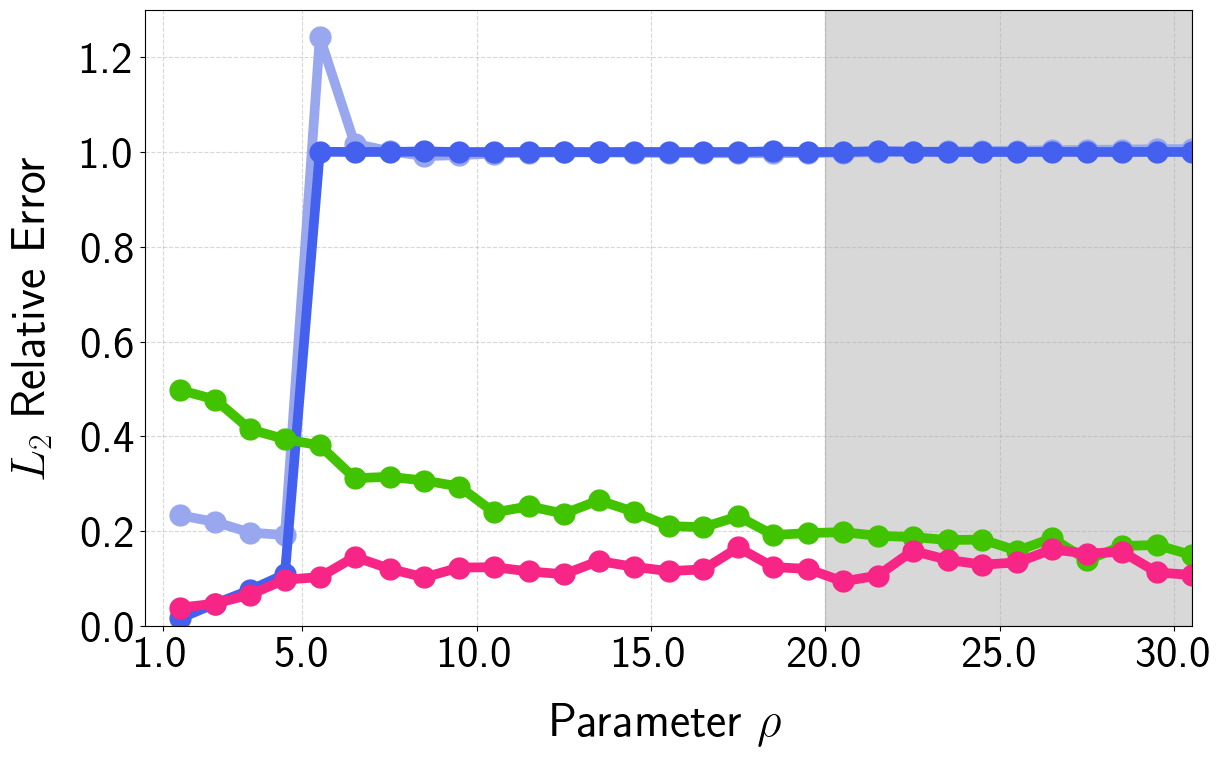}} \hfill
\vspace{-2mm}
\caption{The $L_2$ relative error measured at unseen parameters is presented for (a) convection, (b) diffusion, and (c) reaction, comparing our model with baseline methods. For Hyper-LR-PINN, both fine-tuned and non-fine-tuned results are plotted together. The grey area indicates the region where the model extrapolates the coefficient $\beta, \nu,$ or $\rho$.}
\vskip -0.15in
\label{fig:interpolation}
\end{figure}

From this point, we train our model using only PINN-priors and further explore the base characteristics of SFMs'. In this section, we test our model with unseen parameters at convection, diffusion, and reaction systems. For each system, the model is trained with $[1, 20] \cap \mathbb{Z}$ range coefficients and tested with unseen coefficient $1.5, 2.5, \cdots, 19.5$ which is included in interval $[1, 20]$ and $20.5, 21.5, 22.5, \cdots, 30.5$ which is not in range of $[1, 20]$. 
The $L_2$ relative error measured for each coefficient value is plotted in Figure \ref{fig:interpolation}, along with the baselines Hyper-LR-PINN and the non-fine-tuned P$^2$INN.

Over the trained coefficient range, our model effectively interpolates the coefficients $\beta$, $\nu$, and $\rho$, achieving performance comparable to that seen with known coefficients. Moreover, the model demonstrates stable extrapolation in diffusion and reaction systems. Compared to the baselines, our model significantly outperforms it, particularly in diffusion and reaction systems. This result indicates that the Transformer can effectively learn the PPD of the prior space $\mathcal{D}$, even without observing the complete prior.

\subsubsection{Time Domain Extrapolation for Seen PDE Parameters}
\label{SSS:time_inter_parameter2}

One major limitation of the PINN is an extrapolation at the temporal domain that infer solutions at unknown points. Our model demonstrates extrapolation capability in the 1D convection equation, where the solution exhibits wave-like fluctuations in the inference region. In particular, the model trained with the PINN-prior $\mathcal{D}$ over the coefficient range $\beta \in [1, 20] \cap \mathbb{Z}$ can predict $\beta$ values in ${1.5, 2.5, \cdots, 16.5}$ for equations where the test points $\widetilde{T}$ fall within $t \in (0.6, 1.0]$, even though $\widetilde{D}$ is only distributed within $t \in [0.0, 0.6]$. We then evaluate the relative $L_2$ error and plot for each coefficient $\beta$ with our baselines. Both baselines are not fine-tuned for each test $\beta$ to make a fair comparison with our zero-shot model.


\begin{figure}[t]
    \centering
    \includegraphics[width=1.0\textwidth]{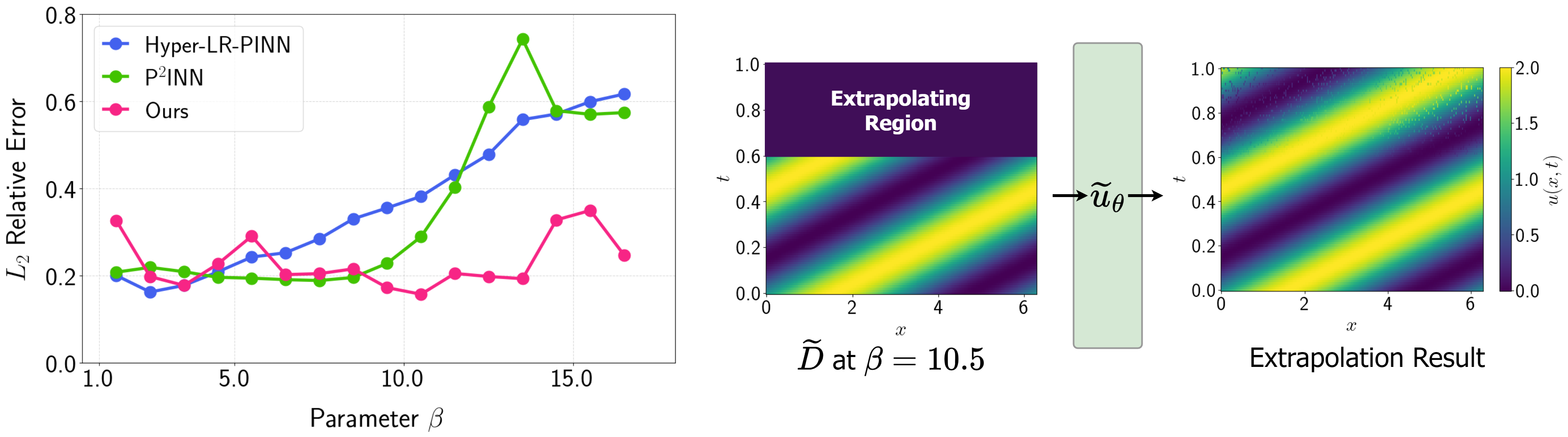}
    \captionsetup{skip=5pt}
    \caption{\textit{(Left)} The $L_2$ relative error is evaluated for each convection coefficient $\beta = 1.5, 2.5, \cdots, 16.5$ as an extrapolation task. \textit{(Right)} The graph illustrates the extrapolation of convection equation with $\beta = 10.5$ at $0.6 \leq t \leq 1.0$.}
    \label{fig:extrapolation}
    \vspace{-17pt}
\end{figure}

As a result, our model demonstrates effective extrapolation capabilities in convection equation (Figure \ref{fig:extrapolation}, \textit{Left}). In addition, our model outperforms both Hyper-LR-PINN and P$^2$INN across most values of $\beta$, while maintaining a stable $L_2$ relative error over a wider range. The diagram in Figure \ref{fig:extrapolation}, \textit{Right} presents the detailed performance at $\beta = 10.5$. This capability emphasizes our model's potential for advancing solutions to PDEs in unknown spatial regions and for enhancing time series predictions.

\subsection{Comprehensive Study with Various Reaction Terms}
\label{SS:comprehensive_study}

In this section, we expand the parameter space to following $\Omega$ using three different reaction terms: Fisher ($f_{1}$), Allen-Cahn ($f_{2}$), and Zeldovich ($f_{3}$),
\begin{equation}
\label{eqs:expanded governing equation}
\begin{aligned}
    u_t = \mathcal{N}(\cdot), \;\;\;\; \mathcal{N}(t, x, u, \alpha) &= -\beta u_x + \nu u_{xx} + \sum_{j=1}^{3} \rho_{j} f_j(u),\\
    f_{1} := u(1-u), \;\; f_{2} := u&(1-u^2), \;\; f_{3} := u^2(1-u),\\
    \Omega = \{ \alpha := (\beta, \nu, & \; \rho_1, \rho_2, \rho_3 ) \}.
\end{aligned}
\end{equation}

To justify the expansion, we train the Transformer with $\beta = 0$, $\nu = 0$, and $\rho_j \in [1, 5] \cap \mathbb{Z}$ for $j = 1, 2, 3$ to evaluate its ICL capability in handling linear combinations of the reaction terms. The model is then tasked with inferring the solutions of the PDEs $u_t = \rho_1 f_1$, $u_t = \rho_2 f_2$, and $u_t = \rho_3 f_3$ to test whether it can generalize to unseen PDEs and accurately distinguish between each component.


\begin{table*}[t]
\caption{The Transformer is trained using a linear combination of Fisher, Allen-Cahn, and Zeldovich reaction terms with a given train parameter range. The $L_2$ absolute and relative errors for inferring each reaction term are then averaged over the given test parameter range. For comparison, the results for the Fisher reaction term tested in Section \ref{SSS:time_inter_numerical} are also included.}
\label{table:reaction_term}
\small
\centering
\begin{tabular}{lcccc}
\toprule
\textbf{Train Parameter Range} & $\rho_1 \in [1, 5] \cap \mathbb{Z}$ & \multicolumn{3}{c}{$\rho_1, \rho_2, \rho_3 \in [1, 5] \cap \mathbb{Z}$} \\
\midrule
\textbf{Test Parameter Range} & $\rho_1 \in [1, 5] \cap \mathbb{Z}$ & $\rho_1 \in [1, 5] \cap \mathbb{Z}$ & $\rho_2 \in [1, 5] \cap \mathbb{Z}$ & $\rho_3 \in [1, 5] \cap \mathbb{Z}$ \\
\midrule
\textbf{$\mathbf{L_2}$ Abs Err.} & 0.0102 & 0.0755 & 0.0381 & 0.0611 \\
\textbf{$\mathbf{L_2}$ Rel Err.} & 0.0154 & 0.1098 & 0.0734 & 0.0830 \\
\bottomrule
\end{tabular}
\end{table*}

According to the result at Table \ref{table:reaction_term}, the $L_2$ absolute and relative errors are comparable to those obtained when trained with $\rho_1 \in [1, 5] \cap \mathbb{Z}$, suggesting the potential for expanding the parameter space. Specifically, the results demonstrate that our model can accurately distinguish between the three different reaction terms, even when trained with their linear combination. Although the model was trained using arbitrary linear combinations of terms commonly found in real-world applications, it is capable of effectively solving PDEs composed of meaningful combinations of these terms during testing. This demonstrates the model’s ability to generalize beyond its training data and infer significant governing relationships from complex systems.

\section{Related Works}

\paragraph{In-context learning} Transformers have shown remarkable ICL abilities across various studies. They can generalize to unseen tasks by emulating Bayesian predictors \citep{panwar2024incontext} and linear models \citep{JMLR:v25:23-1042}, while also efficiently performing Bayesian inference through Prior-Data Fitted Networks (PFNs) \citep{DBLP:journals/corr/abs-2112-10510}. Their robustness extends to learning different classes of functions, such as linear and sparse linear functions, decision trees, and two-layer neural networks even under distribution shifts \citep{NEURIPS2022_c529dba0}. Furthermore, Transformers can adaptively select algorithms based on input sequences, achieving near-optimal performance on tasks like noisy linear models \citep{NEURIPS2023_b2e63e36}. They are also highly effective and fast for tabular data classification \citep{hollmann2022tabpfn}.

\paragraph{Foundation model} Recent studies have advanced in-context operator learning and PDE solving through Transformer-based models. \citet{ye2024pdeformerfoundationmodelonedimensional} introduces PDEformer, a versatile model for solving 1D PDEs with high accuracy and strong performance in inverse problems. In-context operator learning has also been extended to multi-modal frameworks, as seen in \citet{yang2024finetunelanguagemodelsmultimodal}, where ICON-LM integrates natural language and equations to outperform traditional models. Additionally, \citet{YANG2024113379} and \citet{doi:10.1073/pnas.2310142120} demonstrate the generalization capabilities of In-Context Operator Networks (ICON) in solving various PDE-related tasks, highlighting ICON's adaptability and potential for few-shot learning across different differential equation problems. 
Several other studies have addressed the problem of solving various PDEs using a single trained model \citep{hang2024unisolver,herde2024poseidon} 
. However, many of these approaches rely on symbolic PDE information, true or near-true solutions and/or do not support zero-shot in-context learning, making their objectives different from ours.

\section{Conclusion and Limitations}
In this work, we presented MaD-Scientist for scientific machine learning that integrates in-context learning and Bayesian inference for predicting PDE solutions. Our results demonstrate that Transformers, equipped with self-attention and cross-attention mechanisms, can effectively generalize from prior data, even in the presence of noise, and exhibit robust zero-shot learning capabilities. These findings suggest that foundation models in SciML have the potential to follow the development trajectory similar to that of natural language processing foundation models, offering new avenues for further exploration and advancement in the field.

The Transformer used in our study clearly demonstrates the ICL capability, when trained with PINN-based prior. However, it is limited to the CDR equations in our paper. We will consider other types of PDE and more diverse initial and boundary conditions in the future, enhancing its adaptability to real-world scenarios and its role as a foundation model.
\clearpage

{\bf Acknowledgements.}   Sandia National Laboratories is a multimission laboratory managed and operated by National Technology \& Engineering Solutions of Sandia, LLC, a wholly owned subsidiary of Honeywell International Inc., for the U.S. Department of Energy’s National Nuclear Security Administration under contract DE-NA0003525. This paper describes objective technical results and analysis. Any subjective views or opinions that might be expressed in the paper do not necessarily represent the views of the U.S. Department of Energy or the United States Government. This article has been co-authored by an employee of National Technology \& Engineering Solutions of Sandia, LLC under Contract No. DE-NA0003525 with the U.S. Department of Energy (DOE). The employee owns all right, title and interest in and to the article and is solely responsible for its contents. The United States Government retains and the publisher, by accepting the article for publication, acknowledges that the United States Government retains a non-exclusive, paid-up, irrevocable, world-wide license to publish or reproduce the published form of this article or allow others to do so, for United States Government purposes. The DOE will provide public access to these results of federally sponsored research in accordance with the DOE Public Access Plan \textit{https://www.energy.gov/downloads/doe-public-access-plan}. K. Lee acknowledges support from the U.S. National Science Foundation under grant IIS 2338909.

\section*{Ethics Statement}
This research adheres to the ethical standards required for scientific inquiry. We have considered the potential societal impacts of our work and have found no clear negative implications. All experiments were conducted in compliance with relevant laws and ethical guidelines, ensuring the integrity of our findings. We are committed to transparency and reproducibility in our research processes.

\section*{Reproducibility}
We are committed to ensuring the reproducibility of our research. All experimental procedures, data sources, and algorithms used in this study are clearly documented in the paper. The code and datasets will be provided as the supplementary material and be made publicly available upon publication, allowing others to validate our findings and build upon our work.



\bibliography{iclr2025_conference}

\begin{thebibliography}{47}
\providecommand{\natexlab}[1]{#1}
\providecommand{\url}[1]{\texttt{#1}}
\expandafter\ifx\csname urlstyle\endcsname\relax
  \providecommand{\doi}[1]{doi: #1}\else
  \providecommand{\doi}{doi: \begingroup \urlstyle{rm}\Url}\fi

\bibitem[Al-Khaled(2001)]{ALKHALED2001245}
Kamel Al-Khaled.
\newblock Numerical study of fisher's reaction–diffusion equation by the sinc collocation method.
\newblock \emph{Journal of Computational and Applied Mathematics}, 137\penalty0 (2):\penalty0 245--255, 2001.
\newblock ISSN 0377-0427.
\newblock \doi{https://doi.org/10.1016/S0377-0427(01)00356-9}.
\newblock URL \url{https://www.sciencedirect.com/science/article/pii/S0377042701003569}.

\bibitem[Bai et~al.(2023)Bai, Chen, Wang, Xiong, and Mei]{NEURIPS2023_b2e63e36}
Yu~Bai, Fan Chen, Huan Wang, Caiming Xiong, and Song Mei.
\newblock Transformers as statisticians: Provable in-context learning with in-context algorithm selection.
\newblock In A.~Oh, T.~Naumann, A.~Globerson, K.~Saenko, M.~Hardt, and S.~Levine (eds.), \emph{Advances in Neural Information Processing Systems}, volume~36, pp.\  57125--57211. Curran Associates, Inc., 2023.
\newblock URL \url{https://proceedings.neurips.cc/paper_files/paper/2023/file/b2e63e36c57e153b9015fece2352a9f9-Paper-Conference.pdf}.

\bibitem[Beck \& Kurz(2021)Beck and Kurz]{beck2021perspective}
Andrea Beck and Marius Kurz.
\newblock A perspective on machine learning methods in turbulence modeling.
\newblock \emph{GAMM-Mitteilungen}, 44\penalty0 (1):\penalty0 e202100002, 2021.

\bibitem[Blasi \& Walker(2013)Blasi and Walker]{6e9902fe-c806-359d-8aa7-a66426017a1c}
Pierpaolo~De Blasi and Stephen~G. Walker.
\newblock Bayesian asymptotics with misspecified models.
\newblock \emph{Statistica Sinica}, 23\penalty0 (1):\penalty0 169--187, 2013.
\newblock ISSN 10170405, 19968507.
\newblock URL \url{http://www.jstor.org/stable/24310519}.

\bibitem[Bodnar et~al.(2024)Bodnar, Bruinsma, Lucic, Stanley, Brandstetter, Garvan, Riechert, Weyn, Dong, Vaughan, Gupta, Thambiratnam, Archibald, Heider, Welling, Turner, and Perdikaris]{bodnar2024aurora}
Cristian Bodnar, Wessel Bruinsma, Ana Lucic, Megan Stanley, Johannes Brandstetter, Patrick Garvan, Maik Riechert, Jonathan Weyn, Haiyu Dong, Anna Vaughan, Jayesh Gupta, Kit Thambiratnam, Alex Archibald, Elizabeth Heider, Max Welling, Richard Turner, and Paris Perdikaris.
\newblock Aurora: A foundation model of the atmosphere.
\newblock Technical Report MSR-TR-2024-16, Microsoft Research AI for Science, May 2024.
\newblock URL \url{https://www.microsoft.com/en-us/research/publication/aurora-a-foundation-model-of-the-atmosphere/}.

\bibitem[Brown et~al.(2020)Brown, Mann, Ryder, Subbiah, Kaplan, Dhariwal, Neelakantan, Shyam, Sastry, Askell, Agarwal, Herbert-Voss, Krueger, Henighan, Child, Ramesh, Ziegler, Wu, Winter, Hesse, Chen, Sigler, Litwin, Gray, Chess, Clark, Berner, McCandlish, Radford, Sutskever, and Amodei]{NEURIPS2020_1457c0d6}
Tom Brown, Benjamin Mann, Nick Ryder, Melanie Subbiah, Jared~D Kaplan, Prafulla Dhariwal, Arvind Neelakantan, Pranav Shyam, Girish Sastry, Amanda Askell, Sandhini Agarwal, Ariel Herbert-Voss, Gretchen Krueger, Tom Henighan, Rewon Child, Aditya Ramesh, Daniel Ziegler, Jeffrey Wu, Clemens Winter, Chris Hesse, Mark Chen, Eric Sigler, Mateusz Litwin, Scott Gray, Benjamin Chess, Jack Clark, Christopher Berner, Sam McCandlish, Alec Radford, Ilya Sutskever, and Dario Amodei.
\newblock Language models are few-shot learners.
\newblock In H.~Larochelle, M.~Ranzato, R.~Hadsell, M.F. Balcan, and H.~Lin (eds.), \emph{Advances in Neural Information Processing Systems}, volume~33, pp.\  1877--1901. Curran Associates, Inc., 2020.
\newblock URL \url{https://proceedings.neurips.cc/paper_files/paper/2020/file/1457c0d6bfcb4967418bfb8ac142f64a-Paper.pdf}.

\bibitem[Chien et~al.(2012)Chien, Hsu, and Hsiao]{chien2012manufacturing}
Chen-Fu Chien, Chia-Yu Hsu, and Chih-Wei Hsiao.
\newblock Manufacturing intelligence to forecast and reduce semiconductor cycle time.
\newblock \emph{Journal of Intelligent Manufacturing}, 23:\penalty0 2281--2294, 2012.

\bibitem[Cho et~al.(2023)Cho, Lee, Rim, and Park]{NEURIPS2023_24f8dd1b}
Woojin Cho, Kookjin Lee, Donsub Rim, and Noseong Park.
\newblock Hypernetwork-based meta-learning for low-rank physics-informed neural networks.
\newblock In A.~Oh, T.~Naumann, A.~Globerson, K.~Saenko, M.~Hardt, and S.~Levine (eds.), \emph{Advances in Neural Information Processing Systems}, volume~36, pp.\  11219--11231. Curran Associates, Inc., 2023.
\newblock URL \url{https://proceedings.neurips.cc/paper_files/paper/2023/file/24f8dd1b8f154f1ee0d7a59e368eccf3-Paper-Conference.pdf}.

\bibitem[Cho et~al.(2024)Cho, Jo, Lim, Lee, Lee, Hong, and Park]{pmlr-v235-cho24b}
Woojin Cho, Minju Jo, Haksoo Lim, Kookjin Lee, Dongeun Lee, Sanghyun Hong, and Noseong Park.
\newblock Parameterized physics-informed neural networks for parameterized {PDE}s.
\newblock In Ruslan Salakhutdinov, Zico Kolter, Katherine Heller, Adrian Weller, Nuria Oliver, Jonathan Scarlett, and Felix Berkenkamp (eds.), \emph{Proceedings of the 41st International Conference on Machine Learning}, volume 235 of \emph{Proceedings of Machine Learning Research}, pp.\  8510--8533. PMLR, 21--27 Jul 2024.
\newblock URL \url{https://proceedings.mlr.press/v235/cho24b.html}.

\bibitem[Choi et~al.(2024)Choi, Yun, Kim, and Hong]{CHOI2024116678}
Junho Choi, Taehyun Yun, Namjung Kim, and Youngjoon Hong.
\newblock Spectral operator learning for parametric pdes without data reliance.
\newblock \emph{Computer Methods in Applied Mechanics and Engineering}, 420:\penalty0 116678, 2024.
\newblock ISSN 0045-7825.
\newblock \doi{https://doi.org/10.1016/j.cma.2023.116678}.
\newblock URL \url{https://www.sciencedirect.com/science/article/pii/S0045782523008010}.

\bibitem[Chowdhery et~al.(2023)Chowdhery, Narang, Devlin, Bosma, Mishra, Roberts, Barham, Chung, Sutton, Gehrmann, Schuh, Shi, Tsvyashchenko, Maynez, Rao, Barnes, Tay, Shazeer, Prabhakaran, Reif, Du, Hutchinson, Pope, Bradbury, Austin, Isard, Gur-Ari, Yin, Duke, Levskaya, Ghemawat, Dev, Michalewski, Garcia, Misra, Robinson, Fedus, Zhou, Ippolito, Luan, Lim, Zoph, Spiridonov, Sepassi, Dohan, Agrawal, Omernick, Dai, Pillai, Pellat, Lewkowycz, Moreira, Child, Polozov, Lee, Zhou, Wang, Saeta, Diaz, Firat, Catasta, Wei, Meier-Hellstern, Eck, Dean, Petrov, and Fiedel]{JMLR:v24:22-1144}
Aakanksha Chowdhery, Sharan Narang, Jacob Devlin, Maarten Bosma, Gaurav Mishra, Adam Roberts, Paul Barham, Hyung~Won Chung, Charles Sutton, Sebastian Gehrmann, Parker Schuh, Kensen Shi, Sasha Tsvyashchenko, Joshua Maynez, Abhishek Rao, Parker Barnes, Yi~Tay, Noam Shazeer, Vinodkumar Prabhakaran, Emily Reif, Nan Du, Ben Hutchinson, Reiner Pope, James Bradbury, Jacob Austin, Michael Isard, Guy Gur-Ari, Pengcheng Yin, Toju Duke, Anselm Levskaya, Sanjay Ghemawat, Sunipa Dev, Henryk Michalewski, Xavier Garcia, Vedant Misra, Kevin Robinson, Liam Fedus, Denny Zhou, Daphne Ippolito, David Luan, Hyeontaek Lim, Barret Zoph, Alexander Spiridonov, Ryan Sepassi, David Dohan, Shivani Agrawal, Mark Omernick, Andrew~M. Dai, Thanumalayan~Sankaranarayana Pillai, Marie Pellat, Aitor Lewkowycz, Erica Moreira, Rewon Child, Oleksandr Polozov, Katherine Lee, Zongwei Zhou, Xuezhi Wang, Brennan Saeta, Mark Diaz, Orhan Firat, Michele Catasta, Jason Wei, Kathy Meier-Hellstern, Douglas Eck, Jeff Dean, Slav Petrov, and Noah Fiedel.
\newblock Palm: Scaling language modeling with pathways.
\newblock \emph{Journal of Machine Learning Research}, 24\penalty0 (240):\penalty0 1--113, 2023.
\newblock URL \url{http://jmlr.org/papers/v24/22-1144.html}.

\bibitem[Dai et~al.(2023)Dai, Sun, Dong, Hao, Ma, Sui, and Wei]{dai2023why}
Damai Dai, Yutao Sun, Li~Dong, Yaru Hao, Shuming Ma, Zhifang Sui, and Furu Wei.
\newblock Why can {GPT} learn in-context? language models implicitly perform gradient descent as meta-optimizers.
\newblock In \emph{ICLR 2023 Workshop on Mathematical and Empirical Understanding of Foundation Models}, 2023.
\newblock URL \url{https://openreview.net/forum?id=fzbHRjAd8U}.

\bibitem[Finn et~al.(2017)Finn, Abbeel, and Levine]{DBLP:conf/icml/FinnAL17}
Chelsea Finn, Pieter Abbeel, and Sergey Levine.
\newblock Model-agnostic meta-learning for fast adaptation of deep networks.
\newblock In Doina Precup and Yee~Whye Teh (eds.), \emph{Proceedings of the 34th International Conference on Machine Learning, {ICML} 2017, Sydney, NSW, Australia, 6-11 August 2017}, volume~70 of \emph{Proceedings of Machine Learning Research}, pp.\  1126--1135. {PMLR}, 2017.
\newblock URL \url{http://proceedings.mlr.press/v70/finn17a.html}.

\bibitem[Frieder et~al.(2023)Frieder, Pinchetti, , Griffiths, Salvatori, Lukasiewicz, Petersen, and Berner]{NEURIPS2023_58168e8a}
Simon Frieder, Luca Pinchetti, , Ryan-Rhys Griffiths, Tommaso Salvatori, Thomas Lukasiewicz, Philipp Petersen, and Julius Berner.
\newblock Mathematical capabilities of chatgpt.
\newblock In A.~Oh, T.~Naumann, A.~Globerson, K.~Saenko, M.~Hardt, and S.~Levine (eds.), \emph{Advances in Neural Information Processing Systems}, volume~36, pp.\  27699--27744. Curran Associates, Inc., 2023.
\newblock URL \url{https://proceedings.neurips.cc/paper_files/paper/2023/file/58168e8a92994655d6da3939e7cc0918-Paper-Datasets_and_Benchmarks.pdf}.

\bibitem[Garg et~al.(2022)Garg, Tsipras, Liang, and Valiant]{NEURIPS2022_c529dba0}
Shivam Garg, Dimitris Tsipras, Percy~S Liang, and Gregory Valiant.
\newblock What can transformers learn in-context? a case study of simple function classes.
\newblock In S.~Koyejo, S.~Mohamed, A.~Agarwal, D.~Belgrave, K.~Cho, and A.~Oh (eds.), \emph{Advances in Neural Information Processing Systems}, volume~35, pp.\  30583--30598. Curran Associates, Inc., 2022.
\newblock URL \url{https://proceedings.neurips.cc/paper_files/paper/2022/file/c529dba08a146ea8d6cf715ae8930cbe-Paper-Conference.pdf}.

\bibitem[Gruver et~al.(2023)Gruver, Finzi, Qiu, and Wilson]{NEURIPS2023_3eb7ca52}
Nate Gruver, Marc Finzi, Shikai Qiu, and Andrew~G Wilson.
\newblock Large language models are zero-shot time series forecasters.
\newblock In A.~Oh, T.~Naumann, A.~Globerson, K.~Saenko, M.~Hardt, and S.~Levine (eds.), \emph{Advances in Neural Information Processing Systems}, volume~36, pp.\  19622--19635. Curran Associates, Inc., 2023.
\newblock URL \url{https://proceedings.neurips.cc/paper_files/paper/2023/file/3eb7ca52e8207697361b2c0fb3926511-Paper-Conference.pdf}.

\bibitem[Hang et~al.(2024)Hang, Ma, Wu, Wang, and Long]{hang2024unisolver}
Zhou Hang, Yuezhou Ma, Haixu Wu, Haowen Wang, and Mingsheng Long.
\newblock Unisolver: Pde-conditional transformers are universal pde solvers.
\newblock \emph{arXiv preprint arXiv:2405.17527}, 2024.

\bibitem[Herde et~al.(2024)Herde, Raoni{\'c}, Rohner, K{\"a}ppeli, Molinaro, de~B{\'e}zenac, and Mishra]{herde2024poseidon}
Maximilian Herde, Bogdan Raoni{\'c}, Tobias Rohner, Roger K{\"a}ppeli, Roberto Molinaro, Emmanuel de~B{\'e}zenac, and Siddhartha Mishra.
\newblock Poseidon: Efficient foundation models for pdes.
\newblock \emph{arXiv preprint arXiv:2405.19101}, 2024.

\bibitem[Hollmann et~al.(2022)Hollmann, M{\"u}ller, Eggensperger, and Hutter]{hollmann2022tabpfn}
Noah Hollmann, Samuel M{\"u}ller, Katharina Eggensperger, and Frank Hutter.
\newblock Tab{PFN}: A transformer that solves small tabular classification problems in a second.
\newblock In \emph{NeurIPS 2022 First Table Representation Workshop}, 2022.
\newblock URL \url{https://openreview.net/forum?id=eu9fVjVasr4}.

\bibitem[Kaplan et~al.(2020)Kaplan, McCandlish, Henighan, Brown, Chess, Child, Gray, Radford, Wu, and Amodei]{kaplan2020scalinglawsneurallanguage}
Jared Kaplan, Sam McCandlish, Tom Henighan, Tom~B Brown, Benjamin Chess, Rewon Child, Scott Gray, Alec Radford, Jeffrey Wu, and Dario Amodei.
\newblock Scaling laws for neural language models.
\newblock \emph{arXiv preprint arXiv:2001.08361}, 2020.

\bibitem[Kim et~al.(2023)Kim, Lee, and Hong]{doi:10.1021/acsmaterialslett.2c01096}
Namjung Kim, Dongseok Lee, and Youngjoon Hong.
\newblock Data-efficient deep generative model with discrete latent representation for high-fidelity digital materials.
\newblock \emph{ACS Materials Letters}, 5\penalty0 (3):\penalty0 730--737, 2023.
\newblock \doi{10.1021/acsmaterialslett.2c01096}.
\newblock URL \url{https://doi.org/10.1021/acsmaterialslett.2c01096}.

\bibitem[Kim et~al.(2024)Kim, Lee, Kim, Lee, and Hong]{kim2024simple}
Namjung Kim, Dongseok Lee, Chanyoung Kim, Dosung Lee, and Youngjoon Hong.
\newblock Simple arithmetic operation in latent space can generate a novel three dimensional graph metamaterials.
\newblock \emph{arXiv preprint arXiv:2404.06671}, 2024.

\bibitem[Krishnapriyan et~al.(2021)Krishnapriyan, Gholami, Zhe, Kirby, and Mahoney]{krishnapriyan2021characterizing}
Aditi Krishnapriyan, Amir Gholami, Shandian Zhe, Robert Kirby, and Michael~W Mahoney.
\newblock Characterizing possible failure modes in physics-informed neural networks.
\newblock \emph{Advances in Neural Information Processing Systems}, 34:\penalty0 26548--26560, 2021.

\bibitem[Lee \& Cant(2024)Lee and Cant]{lee2024grid}
Chin~Yik Lee and Stewart Cant.
\newblock A grid-induced and physics-informed machine learning cfd framework for turbulent flows.
\newblock \emph{Flow, Turbulence and Combustion}, 112\penalty0 (2):\penalty0 407--442, 2024.

\bibitem[Moor et~al.(2023)Moor, Banerjee, Abad, Krumholz, Leskovec, Topol, and Rajpurkar]{Moor2023}
Michael Moor, Oishi Banerjee, Zahra Shakeri~Hossein Abad, Harlan~M. Krumholz, Jure Leskovec, Eric~J. Topol, and Pranav Rajpurkar.
\newblock Foundation models for generalist medical artificial intelligence.
\newblock \emph{Nature}, 616\penalty0 (7956):\penalty0 259–265, April 2023.
\newblock ISSN 1476-4687.
\newblock \doi{10.1038/s41586-023-05881-4}.
\newblock URL \url{http://dx.doi.org/10.1038/s41586-023-05881-4}.

\bibitem[Müller et~al.(2021)Müller, Hollmann, Pineda-Arango, Grabocka, and Hutter]{DBLP:journals/corr/abs-2112-10510}
Samuel Müller, Noah Hollmann, Sebastian Pineda-Arango, Josif Grabocka, and Frank Hutter.
\newblock Transformers can do bayesian inference.
\newblock \emph{CoRR}, abs/2112.10510, 2021.
\newblock URL \url{https://arxiv.org/abs/2112.10510}.

\bibitem[Nagler(2023)]{nagler2023statisticalfoundationspriordatafitted}
Thomas Nagler.
\newblock Statistical foundations of prior-data fitted networks.
\newblock In \emph{International Conference on Machine Learning}, pp.\  25660--25676. PMLR, 2023.

\bibitem[Nicolaou et~al.(2023)Nicolaou, Huo, Chen, Brunton, and Kutz]{nicolaou2023data}
Zachary~G Nicolaou, Guanyu Huo, Yihui Chen, Steven~L Brunton, and J~Nathan Kutz.
\newblock Data-driven discovery and extrapolation of parameterized pattern-forming dynamics.
\newblock \emph{Physical Review Research}, 5\penalty0 (4):\penalty0 L042017, 2023.

\bibitem[Panwar et~al.(2024)Panwar, Ahuja, and Goyal]{panwar2024incontext}
Madhur Panwar, Kabir Ahuja, and Navin Goyal.
\newblock In-context learning through the bayesian prism.
\newblock In \emph{The Twelfth International Conference on Learning Representations}, 2024.
\newblock URL \url{https://openreview.net/forum?id=HX5ujdsSon}.

\bibitem[Quirk \& Serda(2001)Quirk and Serda]{quirk2001semiconductor}
Michael Quirk and Julian Serda.
\newblock \emph{Semiconductor manufacturing technology}, volume~1.
\newblock Prentice Hall Upper Saddle River, NJ, 2001.

\bibitem[Radford et~al.(2019)Radford, Wu, Child, Luan, Amodei, Sutskever, et~al.]{radford_language_2019}
Alec Radford, Jeffrey Wu, Rewon Child, David Luan, Dario Amodei, Ilya Sutskever, et~al.
\newblock Language models are unsupervised multitask learners.
\newblock \emph{OpenAI blog}, 1\penalty0 (8):\penalty0 9, 2019.

\bibitem[Raissi et~al.(2019)Raissi, Perdikaris, and Karniadakis]{RAISSI2019686}
M.~Raissi, P.~Perdikaris, and G.E. Karniadakis.
\newblock Physics-informed neural networks: A deep learning framework for solving forward and inverse problems involving nonlinear partial differential equations.
\newblock \emph{Journal of Computational Physics}, 378:\penalty0 686--707, 2019.
\newblock ISSN 0021-9991.
\newblock \doi{https://doi.org/10.1016/j.jcp.2018.10.045}.
\newblock URL \url{https://www.sciencedirect.com/science/article/pii/S0021999118307125}.

\bibitem[Rouf et~al.(2021)Rouf, Malik, Arif, Sharma, Singh, Aich, and Kim]{rouf2021stock}
Nusrat Rouf, Majid~Bashir Malik, Tasleem Arif, Sparsh Sharma, Saurabh Singh, Satyabrata Aich, and Hee-Cheol Kim.
\newblock Stock market prediction using machine learning techniques: a decade survey on methodologies, recent developments, and future directions.
\newblock \emph{Electronics}, 10\penalty0 (21):\penalty0 2717, 2021.

\bibitem[Subramanian et~al.(2023)Subramanian, Harrington, Keutzer, Bhimji, Morozov, Mahoney, and Gholami]{subramanian2023towards}
Shashank Subramanian, Peter Harrington, Kurt Keutzer, Wahid Bhimji, Dmitriy Morozov, Michael~W. Mahoney, and Amir Gholami.
\newblock Towards foundation models for scientific machine learning: Characterizing scaling and transfer behavior.
\newblock In \emph{Thirty-seventh Conference on Neural Information Processing Systems}, 2023.
\newblock URL \url{https://openreview.net/forum?id=zANxvzflMl}.

\bibitem[Touvron et~al.(2023)Touvron, Lavril, Izacard, Martinet, Lachaux, Lacroix, Rozi{\`e}re, Goyal, Hambro, Azhar, et~al.]{touvron2023llamaopenefficientfoundation}
Hugo Touvron, Thibaut Lavril, Gautier Izacard, Xavier Martinet, Marie-Anne Lachaux, Timoth{\'e}e Lacroix, Baptiste Rozi{\`e}re, Naman Goyal, Eric Hambro, Faisal Azhar, et~al.
\newblock Llama: Open and efficient foundation language models.
\newblock \emph{arXiv preprint arXiv:2302.13971}, 2023.

\bibitem[Walker(2003)]{walker2003sufficient}
Stephen Walker.
\newblock On sufficient conditions for bayesian consistency.
\newblock \emph{Biometrika}, 90\penalty0 (2):\penalty0 482--488, 2003.

\bibitem[Walker(2004{\natexlab{a}})]{walker2004new}
Stephen Walker.
\newblock New approaches to bayesian consistency.
\newblock \emph{The Annals of Statistics}, 32\penalty0 (5), October 2004{\natexlab{a}}.
\newblock ISSN 0090-5364.
\newblock \doi{10.1214/009053604000000409}.
\newblock URL \url{http://dx.doi.org/10.1214/009053604000000409}.

\bibitem[Walker(2004{\natexlab{b}})]{walker2004modern}
Stephen~G Walker.
\newblock Modern bayesian asymptotics.
\newblock \emph{Statistical Science}, pp.\  111--117, 2004{\natexlab{b}}.

\bibitem[Willard et~al.(2022)Willard, Jia, Xu, Steinbach, and Kumar]{10.1145/3514228}
Jared Willard, Xiaowei Jia, Shaoming Xu, Michael Steinbach, and Vipin Kumar.
\newblock Integrating scientific knowledge with machine learning for engineering and environmental systems.
\newblock \emph{ACM Comput. Surv.}, 55\penalty0 (4), nov 2022.
\newblock ISSN 0360-0300.
\newblock \doi{10.1145/3514228}.
\newblock URL \url{https://doi.org/10.1145/3514228}.

\bibitem[Xie et~al.(2024)Xie, Chen, Chen, Peng, Hu, Lin, Peng, Huang, Zhang, Keloth, et~al.]{xie2024llamafoundationlargelanguage}
Qianqian Xie, Qingyu Chen, Aokun Chen, Cheng Peng, Yan Hu, Fongci Lin, Xueqing Peng, Jimin Huang, Jeffrey Zhang, Vipina Keloth, et~al.
\newblock Me llama: Foundation large language models for medical applications.
\newblock \emph{arXiv preprint arXiv:2402.12749}, 2024.

\bibitem[Xu et~al.(2024)Xu, Shi, Wei, Mu, Li, and Liang]{xu2024towards}
Zhuoyan Xu, Zhenmei Shi, Junyi Wei, Fangzhou Mu, Yin Li, and Yingyu Liang.
\newblock Towards few-shot adaptation of foundation models via multitask finetuning.
\newblock In \emph{The Twelfth International Conference on Learning Representations}, 2024.
\newblock URL \url{https://openreview.net/forum?id=1jbh2e0b2K}.

\bibitem[Yang et~al.(2023{\natexlab{a}})Yang, Tian, Tan, Huang, Liu, Chang, Shi, Zhao, Bian, Wu, et~al.]{yang2024uniaudio}
Dongchao Yang, Jinchuan Tian, Xu~Tan, Rongjie Huang, Songxiang Liu, Xuankai Chang, Jiatong Shi, Sheng Zhao, Jiang Bian, Xixin Wu, et~al.
\newblock Uniaudio: An audio foundation model toward universal audio generation.
\newblock \emph{arXiv preprint arXiv:2310.00704}, 2023{\natexlab{a}}.

\bibitem[Yang \& Osher(2024)Yang and Osher]{YANG2024113379}
Liu Yang and Stanley~J. Osher.
\newblock Pde generalization of in-context operator networks: A study on 1d scalar nonlinear conservation laws.
\newblock \emph{Journal of Computational Physics}, pp.\  113379, 2024.
\newblock ISSN 0021-9991.
\newblock \doi{https://doi.org/10.1016/j.jcp.2024.113379}.
\newblock URL \url{https://www.sciencedirect.com/science/article/pii/S0021999124006272}.

\bibitem[Yang et~al.(2023{\natexlab{b}})Yang, Liu, Meng, and Osher]{doi:10.1073/pnas.2310142120}
Liu Yang, Siting Liu, Tingwei Meng, and Stanley~J. Osher.
\newblock In-context operator learning with data prompts for differential equation problems.
\newblock \emph{Proceedings of the National Academy of Sciences}, 120\penalty0 (39):\penalty0 e2310142120, 2023{\natexlab{b}}.
\newblock \doi{10.1073/pnas.2310142120}.
\newblock URL \url{https://www.pnas.org/doi/abs/10.1073/pnas.2310142120}.

\bibitem[Yang et~al.(2023{\natexlab{c}})Yang, Liu, and Osher]{yang2024finetunelanguagemodelsmultimodal}
Liu Yang, Siting Liu, and Stanley~J Osher.
\newblock Fine-tune language models as multi-modal differential equation solvers.
\newblock \emph{arXiv preprint arXiv:2308.05061}, 2023{\natexlab{c}}.

\bibitem[Ye et~al.(2024)Ye, Huang, Chen, Liu, Wang, and Dong]{ye2024pdeformerfoundationmodelonedimensional}
Zhanhong Ye, Xiang Huang, Leheng Chen, Hongsheng Liu, Zidong Wang, and Bin Dong.
\newblock Pdeformer: Towards a foundation model for one-dimensional partial differential equations.
\newblock \emph{arXiv preprint arXiv:2402.12652}, 2024.

\bibitem[Zhang et~al.(2024)Zhang, Frei, and Bartlett]{JMLR:v25:23-1042}
Ruiqi Zhang, Spencer Frei, and Peter~L. Bartlett.
\newblock Trained transformers learn linear models in-context.
\newblock \emph{Journal of Machine Learning Research}, 25\penalty0 (49):\penalty0 1--55, 2024.
\newblock URL \url{http://jmlr.org/papers/v25/23-1042.html}.

\end{thebibliography}
\bibliographystyle{iclr2025_conference}

\newpage

\appendix
\section{The Proof of Theorem 2.1}\label{a:proof}
\begin{proof}
For any \( n, \epsilon \), we derive that
\begin{align*}
\mathbb{E}_{x} \left[ H \left( p_{\hat{\theta}}(\cdot \mid x, D_n), \pi(\cdot \mid x) \right) \right] 
&\leq \mathbb{E}_{x} \left[ H \left( p_{\hat{\theta}}(\cdot \mid x, D_n), \pi(\cdot \mid x, D_n) \right) \right]\tag{1}\label{sensi_1}\\
&\quad + \mathbb{E}_{x} \left[ H \left( \pi(\cdot \mid x, D_n), \pi(\cdot \mid x) \right) \right] \\
&\leq \sqrt{\frac{1}{2} \mathbb{E}_{x} \left[ KL \left( p_{\hat{\theta}}(\cdot \mid x, D_n), \pi(\cdot \mid x, D_n) \right) \right]}\tag{2}\label{sensi_2} \\
&\quad + \mathbb{E}_{x} \left[ H \left( \pi(\cdot \mid x, D_n), \pi(\cdot \mid x) \right) \right]\\
&\leq \sqrt{\frac{\epsilon}{2}} + \mathbb{E}_{x} \left[ 1 - \int_{\mathcal{Y}} \sqrt{\int q(y \mid x) \pi(y \mid x) \mathrm{d}\Pi^n(q)} \mathrm{d}y \right]^{1/2} \tag{3}\label{sensi_3} \\
&\leq \sqrt{\frac{\epsilon}{2}} +\left[ 1 -\int_{\mathcal{X}}\pi(x) \int_{\mathcal{Y}} \frac{1}{\pi(x)}\sqrt{\int q(y, x) \pi(y, x) \mathrm{d}\Pi^n(q)} \mathrm{d}y \mathrm{d}x\right]^{1/2}\\
&\leq \sqrt{\frac{\epsilon}{2}} +  \left[ 1 - \int_{\mathcal{X}}\int_{\mathcal{Y}} \int \sqrt{q(y, x) \pi(y , x)} \mathrm{d}\Pi^n(q) \mathrm{d}y \mathrm{d}x \right]^{1/2} \\
&= \sqrt{\frac{\epsilon}{2}} + 
\left[ \int  H \left( q, \pi \right)^2 \mathrm{d}\Pi^n(q) \right]^{1/2} \\
&\leq \sqrt{\frac{\epsilon}{2}} +  \left[\int H \left( q, \pi\right) d\Pi^n(q) \right]^{1/2} \\
&= \sqrt{\frac{\epsilon}{2}} +  \left[ \int_{\{ q : H(\pi, q) > \epsilon \}} H \left( q, \pi \right) \mathrm{d}\Pi^n(q) \right]^{1/2}\\ 
&\quad + \left[ \int_{\{ q : H(\pi, q) \leq \epsilon \}} H \left( q, \pi \right) \mathrm{d}\Pi^n(q) \right]^{1/2} \tag{4}\label{sensi_4} \\
&= \sqrt{\frac{\epsilon}{2}} + (\Pi^n(\{ q : H(\pi, q) > \epsilon \}) + \epsilon)^{1/2} \rightarrow \sqrt{\frac{\epsilon}{2}} + \sqrt{\epsilon} \quad \text{a.s.}  \tag{5}\label{sensi_5}
\end{align*}

The first inequality (\ref{sensi_1}) is derived from the triangle inequality for the Hellinger distance, which states that for any intermediate distribution \( q(\cdot \mid x, D_n) \), we have
\[
H \left( p_{\hat{\theta}}(\cdot \mid x, D_n), \pi(\cdot \mid x) \right) \leq H \left( p_{\hat{\theta}}(\cdot \mid x, D_n), q(\cdot \mid x, D_n) \right) + H \left( q(\cdot \mid x, D_n), \pi(\cdot \mid x) \right).
\]

The second inequality (\ref{sensi_2}) uses the fact that the Hellinger distance \( H(p, q) \) is bounded above by the square root of the KL divergence \( KL(p \parallel q) \), such that
\[
H(p, q)^2 \leq \frac{1}{2} KL(p \parallel q).
\]
Thus, we can bound the Hellinger distance by the KL divergence. In the third inequality (\ref{sensi_3}), we make use of assumption
\[
\mathbb{E}_{x} \left[ KL \left( p_{\hat{\theta}}(\cdot \mid x, D_n), \pi(\cdot \mid x, D_n) \right) \right] < \epsilon,
\]
and utilize the definition of the Hellinger distance.
In (\ref{sensi_4}), we partition the domain into two regions--one where the Hellinger distance \( H(\pi, q) \) exceeds \( \epsilon \) and another where it is less than or equal to  \( \epsilon \)--and use this partitioning to demonstrate the inequality.

Finally, in (\ref{sensi_5}), by posterior consistency, the region where the Hellinger distance is greater than \( \epsilon \) vanishes as $n \rightarrow \infty$ such that 
\[
\Pi^n \left\{ q : H(\pi, q) > \epsilon \right\} \to 0 \quad \text{almost surely}.
\]
Since $\epsilon$ is arbitrary, we can conclude that \[
\mathbb{E}_{x} \left[ H \left( p_{\hat{\theta}}(\cdot \mid x, D_n), \pi(\cdot \mid x) \right) \right] \xrightarrow{n \to \infty} 0 \quad \text{almost surely}.
\]

\end{proof}

\section{Environments} \label{a:environment}

We conducted the experiments using Python 3.8.19, PyTorch 2.4.0+cu121, scikit-learn 1.3.2, NumPy 1.24.4, and pandas 2.0.3, with CUDA 12.1, NVIDIA Driver 535.183.01, and an NVIDIA RTX A6000. Additionally, we used CUDA 12.4, NVIDIA Driver 550.67, and either an NVIDIA GeForce RTX 3090 or an NVIDIA TITAN RTX.

\section{Additional Comparison Between Baselines} \label{a:Comparison}

In addition to the comparison between baselines in \ref{table:baseline_comparison}, the additional comparisons between baselines is shown below. For Hyper-LR-PINN, the results are estimated solely for Phase 1 and the number of model parameters estimated refers specifically to the model used in sections~\ref{SS:focused_study}.


\begin{table}[htbp]
\caption{Additional comparisons between baselines.}
\label{table:TableA}
\small
\centering
\makebox[\textwidth][c]{
\resizebox{\textwidth}{!}{%
\begin{tabular}{l >{\centering\arraybackslash}p{3cm} >{\centering\arraybackslash}p{3cm} >{\centering\arraybackslash}p{3cm}}
\toprule
\textbf{Properties} & \textbf{Hyper-LR-PINN} & \textbf{P$^2$INN} & \textbf{Ours} \\
\midrule
Number of model parameters & 28,151 & 76,851 & \textbf{21,697} \\
GPU memory usage & 1,175MB & 1,108MB & \textbf{998MB} \\
Training time per epoch & 1.8613s & 0.9640s & \textbf{0.1187s} \\
\bottomrule
\end{tabular}
}}
\end{table}

\section{Datasets} \label{a:datasets}

The specific information about PDE types used in the study is following. $n$ represent the maximum values for each parameter $\beta, \nu, \rho_1, \rho_2$ and $\rho_3$'s range, which are set to 5, 10, and 20 in our study.


\begin{table}[htbp]
\caption{PDEs used in our study and corresponding dataset information for each section. "c-d-r" means "convection-diffusion-reaction".}
\label{table:TableB}
\small
\centering
\makebox[\textwidth][c]{
\resizebox{\textwidth}{!}{%
\begin{tabular}{>{\centering\arraybackslash}p{1cm} > {\arraybackslash}p{2.7cm} > {\arraybackslash}p{4.2cm} >{\centering\arraybackslash}p{3.3cm} >{\centering\arraybackslash}p{2.5cm}}
\toprule
\textbf{Section} & \textbf{Equation Type} & \textbf{Equation} & \textbf{Parameter Range} & \textbf{Number of Datasets} \\
\midrule

\multirow{6}{*}{\textbf{~\ref{SS:focused_study}}} &

convection & $u_t = -\beta u_x$ & $\beta \in [1, n] \cap \mathbb{Z}$ & $n$\\
& diffusion & $u_t = \nu u_{xx}$ &$\nu \in [1, n] \cap \mathbb{Z}$ & $n$ \\
& reaction & $u_t = \rho_1 u(1-u)$ & $\rho_1 \in [1, n] \cap \mathbb{Z}$ & $n$ \\
& convection-diffusion & $u_t = -\beta u_x + \nu u_{xx}$ & $\beta, \nu \in [1, n] \cap \mathbb{Z}$ & $n^2$ \\
& reaction-diffusion & $u_t = \nu u_{xx} + \rho_1 u(1-u)$ & $\nu, \rho_1 \in [1, n] \cap \mathbb{Z}$ & $n^2$ \\
& c-d-r & $u_t = - \beta u_x + \nu u_{xx} + \rho_1 u(1-u)$ & $\beta, \nu, \rho_1 \in [1, n] \cap \mathbb{Z}$ & $n^3$ \\

\midrule

\multirow{1}{*}{\textbf{~\ref{SS:comprehensive_study}}} &

Table \ref{table:reaction_term} & $u_t = \rho_1 u(1-u) + \rho_2 u(1-u^2) + \rho_3 u^2(1-u)$ & $\rho_1, \rho_2, \rho_3 \in [1, n] \cap \mathbb{Z}$ & $n^3$ \\

\bottomrule
\end{tabular}
}}
\end{table}

For each specific PDE, we collect 256 initial points, 1,000 collocation points, 100 boundary points, and 1,000 test points. For each 1,000 collocation points, we sample 800 points for $D \cup T$ and 200 points for $\Tilde{D}$ which are not overlapped. 
During the training phase, 30\% of the data points are designated as $D$, while the remaining 70\% are allocated to $T$.

\section{PINN Used in Prior Generation}
\label{a:PINN}

In this study, we utilize the PINN introduced by \citet{RAISSI2019686} to generate PINN-priors. The loss function employed during the training of the PINN is as follows:
\begin{equation}
\mathcal{L} = \mathcal{L}_u + \mathcal{L}_f + \mathcal{L}_b,
\end{equation}
where $\mathcal{L}_u, \mathcal{L}_f$ and $\mathcal{L}_b$ is defined as
\begin{equation}
\mathcal{L}_u = \frac{1}{N_u} \sum \left( \Tilde{u}(x, 0) - u(x, 0) \right)^2, \;\; \mathcal{L}_f = \frac{1}{N_f} \sum \left( \mathcal{N}(t, x, u, \alpha) \right)^2, \;\; \mathcal{L}_b = \frac{1}{N_b} \sum \left( \Tilde{u}(0, t) - \Tilde{u}(2 \pi, t) \right)^2,
\end{equation}
for $N_u$ points at initial condition, $N_f$ collocation points and $N_b$ boundary points.

\newpage

\section{Training Algorithm}
\label{a: Training Algorithm}

We train the Transformer like following.

\begin{algorithm}[htbp]
\caption{Training a Transformer}
\label{alg:train_transformer}
\begin{algorithmic}[1]
\STATE \textbf{Input:} {A prior dataset $D \cup T$ drawn from prior $p(\mathcal{D})$}
\STATE \textbf{Output:} {A Transformer $\tilde{u}_{\theta}$ which can approximate the PPD}
\STATE Initialize the Transformer $\Tilde{u}_{\theta}$
\FOR{$i=1$ to $n$}
    \STATE Sample $\alpha \in \Omega$ and $D \cup T \subseteq \Tilde{u}(\alpha) \sim p(\mathcal{D})$
    \STATE ($D := \{(x_D^{(i)}, t_D^{(i)})\}_{i=1}^{N_D}$, $T := \{(x_T^{(j)}, t_T^{(j)})\}_{j=1}^{N_T}$)
    \STATE Compute loss $L_{\alpha} = \frac{1}{N_T} \sum_{j=1}^{N_T} \left\{ \Tilde{u}_{\theta}(x_T^{(j)}, t_T^{(j)} | D_n) - \Tilde{u}(x_T^{(j)}, t_T^{(j)})\right\}^2.$
    \STATE Update parameters $\theta$ with an Adam optimizer
\ENDFOR

\end{algorithmic}
\end{algorithm}

\vspace{-3pt}

\section{ICL of Transformers with Noisy Prior}\label{a:noisy}

In order to study the ICL capability of Transformers with noisy prior, we introduce four kinds of prior $\mathcal{D}$ like 

\begin{flushleft}
    \begin{minipage}[t]{0.45\textwidth}
        \textbf{P1} (noiseless) : $p(\mathcal{D}) = p(\mathcal{U})$,
    \end{minipage}
    \begin{minipage}[t]{0.45\textwidth}
        \textbf{P2} (Gaussian noise) : $p(\mathcal{D}) \sim \mathcal{N}(\mathcal{U}, \sigma^2 \mathbf{I})$,
    \end{minipage}
\end{flushleft}

\textbf{P3} (salt-and-pepper noise) : $p(\mathcal{D}) \sim p(s \cdot \mathcal{U})$ where $s = 
\begin{cases}
\min(\mathcal{U}) & \text{with probability } \frac{\gamma}{2}, \\
\max(\mathcal{U}) & \text{with probability } \frac{\gamma}{2}, \\
1 & \text{with probability } 1-\gamma,
\end{cases}$

\textbf{P4} (uniform noise) : $p(\mathcal{D}) \sim p(\mathcal{U} + U(-\epsilon, \epsilon))$ ($U$: uniform distribution).
\\
We sample $D \cup T \sim p(\mathcal{D})$, where $\mathcal{D}$ is a noisy prior, and train the Transformer $\widetilde{u}_{\theta}$. We then test $\widetilde{u}_{\theta}$ with $\widetilde{D} \cup \widetilde{T} \sim p(\mathcal{U})$, demonstrating that the model can predict the true solution even when trained on noisy prior data. The experiment is conducted on reaction and convection-diffusion-reaction equations, which outperform other baselines, under three different noises: the Gaussian noise (\textbf{P2}), the salt-and-pepper noise (\textbf{P3}), and the uniform noise (\textbf{P4}). The standard deviation $\sigma$ of Gaussian noise is set to 1\%, 5\%, and 10\% of the mean value of the ground truth solution. Additionally, for the experiment, the probe $\gamma$ for salt-and-pepper noise and the range $\epsilon$ for uniform noise are also set to 1\%, 5\%, and 10\%.


\begin{table*}[htbp]
\caption{The relative and absolute $L_2$ errors for the reaction and convection-diffusion-reaction systems using the \textbf{P2} prior with varying levels of Gaussian noise $\sigma$, \textbf{P3} prior with varying levels of noise probe $\gamma$, and \textbf{P4} prior with varying levels of noise $\epsilon$ (1\%, 5\%, and 10\%). For a comparison, the result of using \textbf{P1} prior is notated.}
\label{table:noisy prior}

\vskip 0.15in
\begin{center}
\makebox[\textwidth][c]{
\resizebox{\textwidth}{!}{%
\begin{tabular}{lcccccccccr}
\toprule
\multirow{3}{*}{\textbf{System}} & \multirow{3}{*}{\textbf{Prior Type}} & \multicolumn{6}{c}{\textbf{Noisy Prior with a Noise Level}} & \multicolumn{2}{c}{\multirow{2}{*}{\textbf{P1 Prior}}} \\
\cmidrule(lr){3-8}
&& \multicolumn{2}{c}{\textbf{1\% Noise}} & \multicolumn{2}{c}{\textbf{5\% Noise}} & \multicolumn{2}{c}{\textbf{10\% Noise}} & & \\
\cmidrule(lr){3-8}
&& Abs. err & Rel. err & Abs. err & Rel. err & Abs. err & Rel. err & Abs. err & Rel. err \\
\midrule
\multirow{3}{*}{\textbf{Reaction}} 
& \textbf{P2} & 0.0210 & 0.0392 & 0.0213 & 0.0399 & 0.0210 & 0.0392 & \multirow{3}{*}{0.0160} &\multirow{3}{*}{0.0322}\\ 
\cmidrule(lr){2-8}
& \textbf{P3} & 0.0309 & 0.0598 & 0.0286 & 0.0517 & 0.0354 & 0.0619 &&\\ 
\cmidrule(lr){2-8}
& \textbf{P4} & 0.0285 & 0.0568 & 0.0293 & 0.0583 & 0.0306 & 0.0607 &&\\ 
\midrule

\multirow{3}{*}{\parbox[c]{2cm}{\textbf{Convection}\\ \textbf{-Diffusion}\\ \textbf{-Reaction}}} 
& \textbf{P2} & 0.0175 & 0.0296 & 0.0220 & 0.0431 & 0.0235 & 0.0431 & \multirow{3}{*}{0.0159} &\multirow{3}{*}{0.0310}\\ 
\cmidrule(lr){2-8}
& \textbf{P3} & 0.0246 & 0.0459 & 0.0263 & 0.0453 & 0.0267 & 0.0496 &&\\ 
\cmidrule(lr){2-8}
& \textbf{P4} & 0.0210 & 0.0420 & 0.0215 & 0.0422 & 0.0230 & 0.0426 &&\\ 
\bottomrule
\end{tabular}
}}
\end{center}
\vskip -0.1in

\end{table*}

Our model demonstrates robust performance across different types of noise injection as shown in Table \ref{table:noisy prior}. It shows our Transformer can perform ICL with zero-shot learning even if it is trained with inaccurate or noisy prior $D \cup T \sim p(\mathcal{D})$.

\newpage
\section{Experiments at PINN Failure Modes}

Referring to \citet{pmlr-v235-cho24b} and \citet{krishnapriyan2021characterizing}, we test our method on PINN's major failure modes: $\beta \in [30, 40]$ with an initial condition $1+\sin(x)$ and $\rho \in [1, 10]$ with an initial condition $\mathcal{N}\left(\pi, \left(\frac{\pi}{2}\right)^2\right)$. We have trained our model with this range with \textbf{P1} prior and evaluate $L_2$ absolute and relative errors. The following are major results and solution profiles at failure modes. 


\begin{table*}[h]
\caption{The $L_2$ absolute and relative error at PINN failure modes.}
\begin{center}
\makebox[\textwidth][c]{
\scalebox{0.75}{%
\begin{tabular}{lcccccc}
\toprule
\multirow{2}{*}{\textbf{Trained Coefficient Range}} & \multirow{2}{*}{\textbf{Test Coefficient Value}} & \multicolumn{2}{c}{\textbf{$L_2$ Error Type}} & \multicolumn{2}{c}{\textbf{Average Error}} \\
\cmidrule(lr){3-4} \cmidrule(lr){5-6}
&& Abs.err & Rel.err & Abs.err & Rel.err \\
\midrule

\multirow{4}{*}{$\beta \in [30, 40]$} 
& $\beta = 30$ & 0.2483 & 0.2516 & \multirow{4}{*}{0.1280} & \multirow{4}{*}{0.1328} \\
\cmidrule(lr){2-4}
& $\beta = 31$ & 0.1029 & 0.1111 & & \\
\cmidrule(lr){2-4}
& $\beta = 32$ & 0.0803 & 0.0882 & & \\
\cmidrule(lr){2-4}
& $\beta = 33$ & 0.0806 & 0.0801 & & \\
\midrule

\multirow{4}{*}{$\rho \in [1, 10]$} 
& $\rho = 4$  & 0.0071 & 0.0160 & \multirow{4}{*}{0.0048} & \multirow{4}{*}{0.0097} \\
\cmidrule(lr){2-4}
& $\rho = 5$  & 0.0029 & 0.0054 & & \\
\cmidrule(lr){2-4}
& $\rho = 6$  & 0.0033 & 0.0063 & & \\
\cmidrule(lr){2-4}
& $\rho = 7$  & 0.0058 & 0.0112 & & \\
\bottomrule
\end{tabular}
}} 
\end{center}
\vskip -0.1in
\end{table*}

\begin{figure}[htbp]
\label{fig:profiles}
\centering
\subfigure[$\beta = 30$]{\includegraphics[width=0.24\columnwidth]{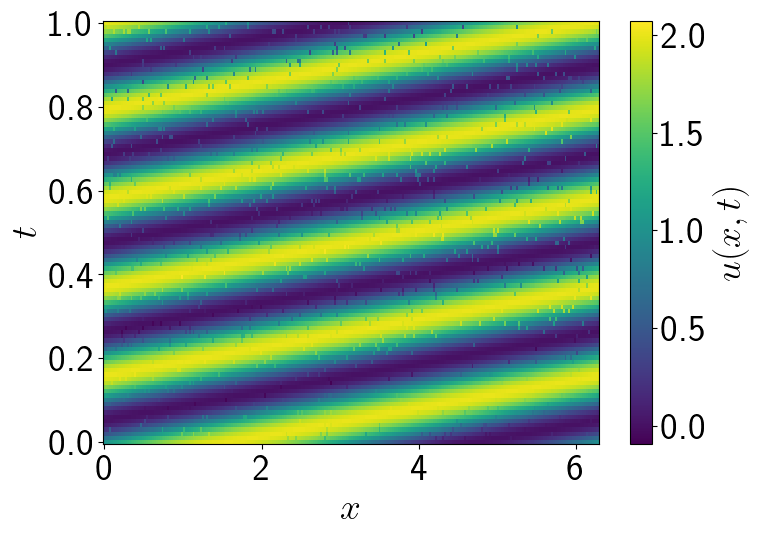}} \hfill
\subfigure[$\beta = 31$]{\includegraphics[width=0.24\columnwidth]{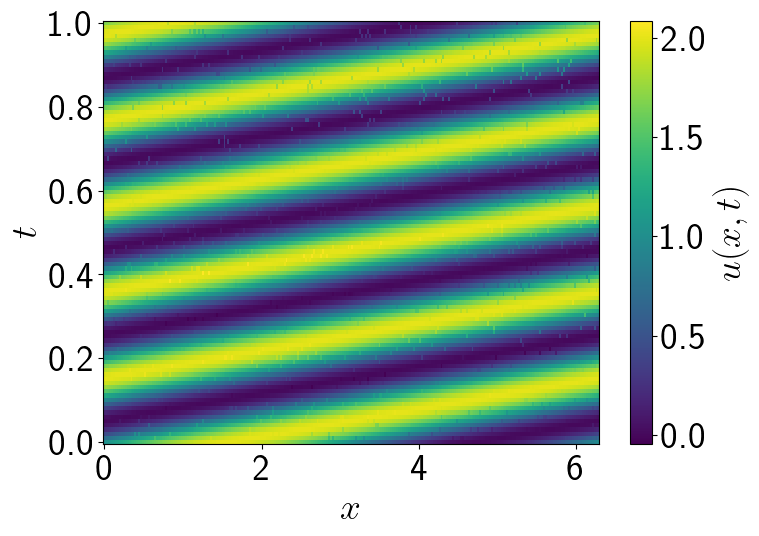}} \hfill
\subfigure[$\beta = 32$]{\includegraphics[width=0.24\columnwidth]{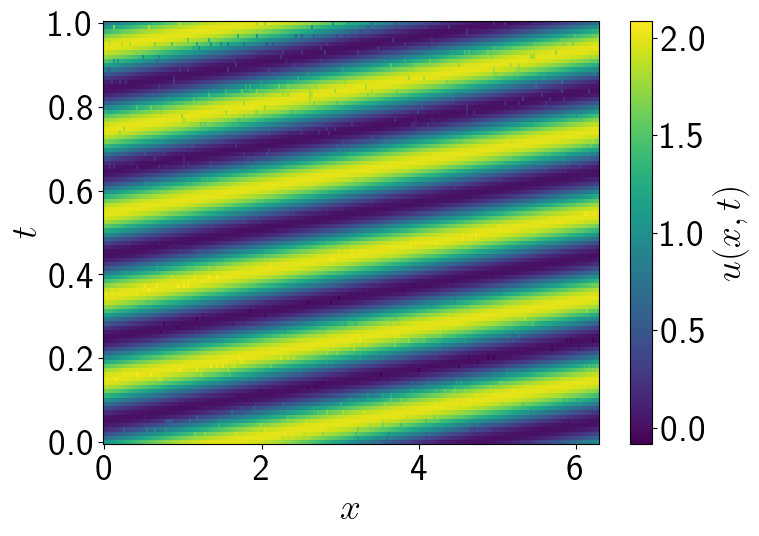}} \hfill
\subfigure[$\beta = 33$]{\includegraphics[width=0.24\columnwidth]{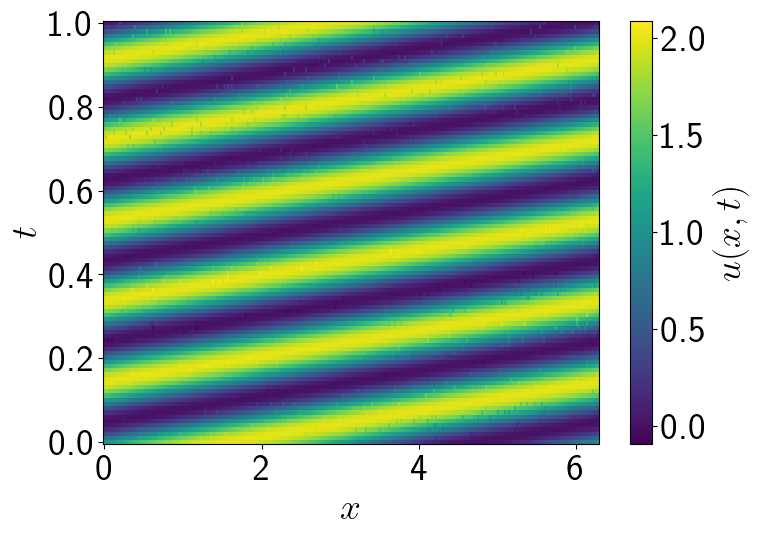}} \hfill
\vspace{-2mm}
\subfigure[$\rho = 4$]{\includegraphics[width=0.24\columnwidth]{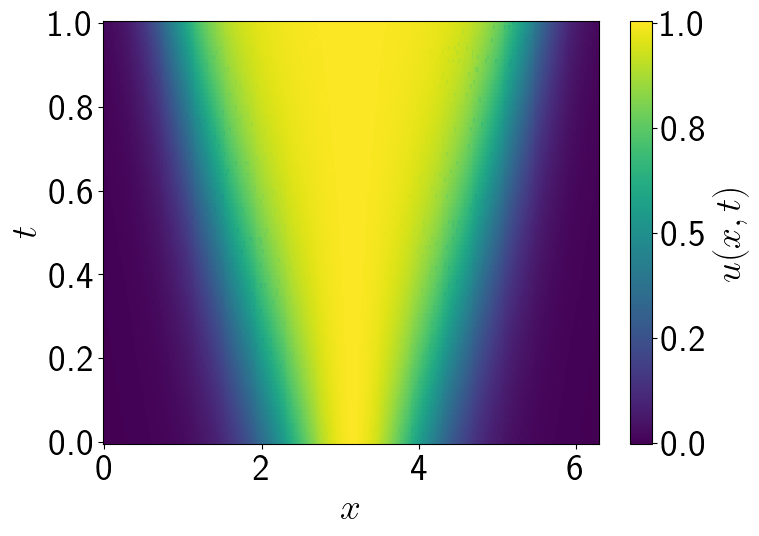}} \hfill
\subfigure[$\rho = 5$]{\includegraphics[width=0.24\columnwidth]{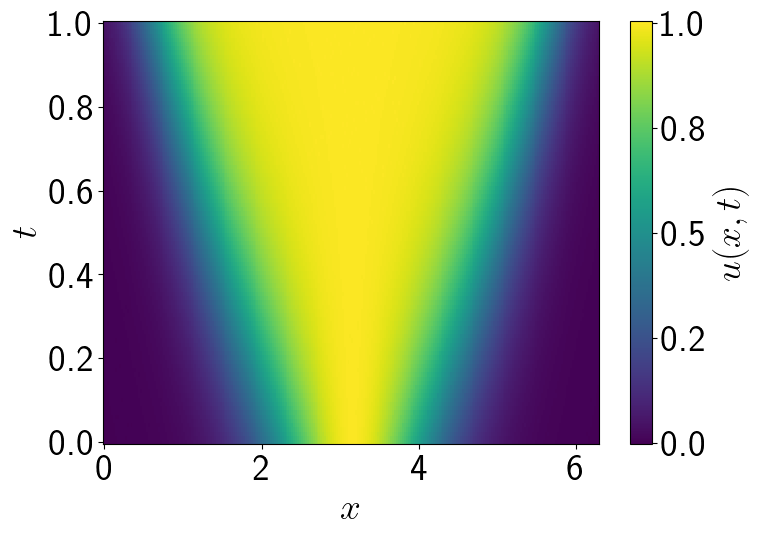}} \hfill
\subfigure[$\rho = 6$]{\includegraphics[width=0.24\columnwidth]{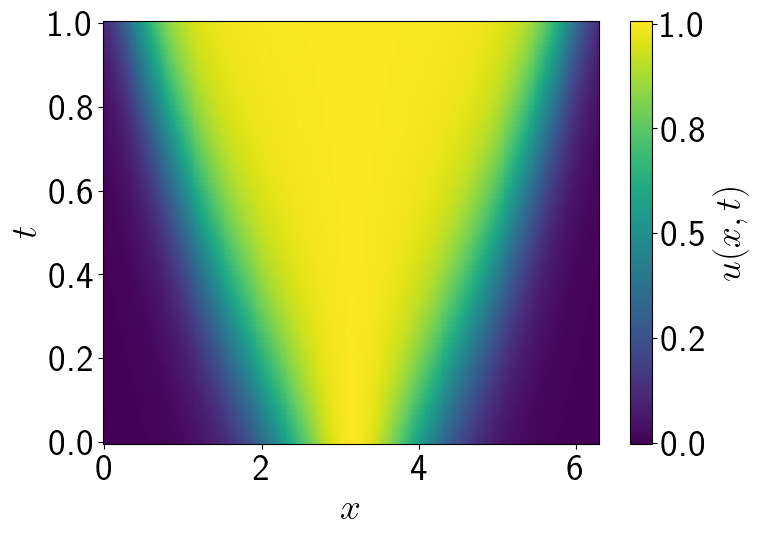}} \hfill
\subfigure[$\rho = 7$]{\includegraphics[width=0.24\columnwidth]{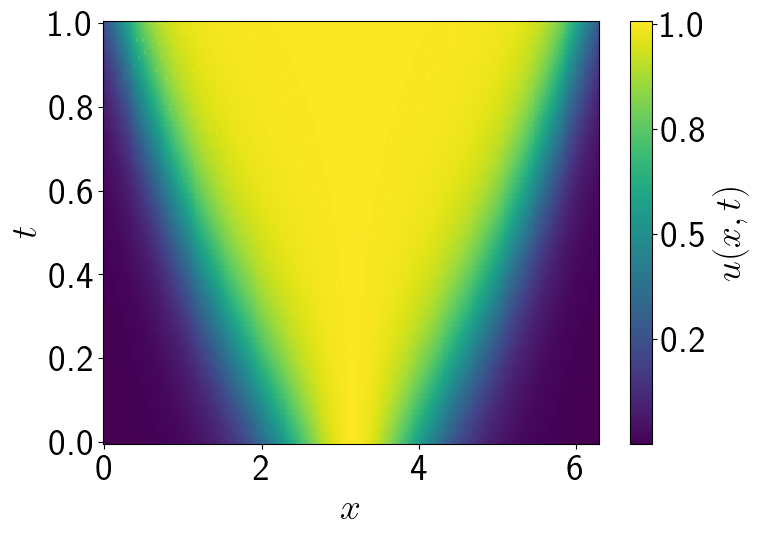}} \hfill

\caption{The solution profiles at PINN failure modes: (a), (b), (c) and (d) for $\beta \in [30, 40]$ with initial condition $1 + \sin(x)$ and (e), (f), (g) and (h) for $\rho \in [1, 10]$ with initial condition $\mathcal{N}\left(\pi, \left(\frac{\pi}{2}\right)^2\right)$. The solution profile is constructed using the union of 1,000 test prediction points and the remaining ground truth points.}
\end{figure}

\section{Hyperparameter List} \label{app:hyperparameter}
The following hyperparameters were employed during the PINN-prior generation and the Transformer training process:
\vspace{-2pt}
\begin{table}[htbp] \label{tab:hyperparameter}
\captionsetup{skip=1.5pt}
\caption{Hyperparameter configuration}
\small
\centering
\resizebox{0.3\paperwidth}{!}{%
\begin{tabular}{lc}
\toprule
\textbf{Hyperparameter Name} & \textbf{Hyperparameter value}\\ 
\midrule
PINN training loss threshold     & $1 \times 10^{-3}$                \\
PINN maximum training epoch & 100                  \\
PINN learning rate          & $1 \times 10^{-2}$                 \\
Transformer layers          & 3                    \\
Transformer hidden size     & 32                   \\
Transformer learning rate   & $1 \times 10^{-5}$  \\
\bottomrule
\end{tabular}%
}
\end{table}
\vspace{-3pt}

\end{document}